# Matching of signal, noise and hardware timescales for filtering and forecasting of correlated noise signals

*Joshua Donald, Alex Gabbitas, Arthur G. T. Coveney, Sergey Savel'ev, Pavel Borisov**

Department of Physics, Loughborough University, Loughborough, LE11 3TU, United Kingdom
E-mail: p.borisov@lboro.ac.uk

Funding:

UK Engineering and Physical Sciences Research Council (EPSRC) studentship 2764909

UK Engineering and Physical Sciences Research Council (EPSRC) research grant EP/T027479/1

## Abstract

Physical reservoir computing exploits the nonlinear dynamics of physical systems to process time-dependent data with greater energy efficiency than conventional machine learning approaches. However, physical reservoirs have fixed intrinsic response timescales, whereas real-world signals combine deterministic and stochastic components across multiple timescales. Here we show, using a nanoporous niobium oxide reservoir, synthetic noisy signals and cryptocurrency-price volatility, that the relationship among noise correlation time, reservoir memory and forecast horizon determines whether correlated noise is filtered or predicted. Noise varying faster than the relevant reservoir memory and forecast horizon is averaged by the reservoir, whereas the temporal structure of slower-varying noise is sufficient for algorithmic forecasting. We introduce the reservoir memory horizon and forecasting regime index to distinguish these operating regimes. These contributions demonstrate that timescale matching can guide the encoding of input time series and development of physical reservoir architectures that filter, analyse and predict stochastic signal components across distinct temporal scales.

## 1. Introduction

Real-world time-dependent signals, such as environmental sensors monitoring[1], medical technology[2], weather or climate change forecasts[3] and air turbulences analysis[4], include a sizeable proportion of noise caused by a variety of sources, including, measurement noise or system-related fluctuations on multiple timescales. On one hand, noise can obscure the deterministic signal component and make the prediction and reconstruction tasks more complicated as a result. Temporally correlated noise on the other hand introduces its own timescales, meaning the corresponding stochastic component of the total signal may itself contain exploitable temporal structure. This so called coloured noise has been highlighted in different applications ranging from physiology (fluctuations in heartbeat, gait and other regulatory signals[5]), neuroimaging (fMRI time series [6]), navigation and geodesy[7,8], climate science (red and fractional noise models[9,10]) to emerging and traditional financial markets (cryptocurrency price series , volatility clustering[11]).

When a machine learning (ML) algorithm is supposed to infer, reconstruct and forecast temporal signals which include noise, the relevant task may vary depending on the corresponding correlation times: certain components of the input signal should be reconstructed; rapidly varying components could be filtered whereas slowly varying components with longer correlation times may remain predictable for a given forecast horizon. Although some machine learning methods were developed for filtering and denoising, or forecasting noisy data[12–14], less attention has been given to distinguishing the computational role of each signal noise component based on its timescale position relative to the prediction horizon.

In comparison to the software-based neural network solutions, physical reservoir computing (RC) realised through physical hardware, with devices such as memristors, delivers higher energy efficiency and demonstrates naturally rich and complex signal dynamics. RC systems[15–17] in general, whilst being part of conventional recurrent artificial neural networks, possess the distinct advantage of requiring only their readout layer to be trained[18], greatly reducing computational costs whilst maintaining similar performance metrics for inference tasks related to temporal signals. RC uses a high-dimensional, non-linear hidden layer (reservoir) with recurrent intrinsic connections and fading memory which performs the transformation of the input to an output layer that can be classified using linear regression.

Volatile, oxide-based memristors are particularly promising as part of the physical RC substrate for a reservoir as they possess non-linearity, short-term memory, are compatible with existing electronics architecture on a compact footprint, and thus can transform time-dependent input signals without a fully software-based network. However, the dynamics of the physical reservoir is governed by the physical processes and characteristics of the device, and hence those cannot be easily reprogrammed as software-based solutions.

Compared to the standard memristor approaches, the in-materia RC systems operate using the dynamics of structurally disordered materials. The disorder and variability present in other devices are used as a computational feature and a resource to be exploited instead. Previous studies demonstrated successful applications of physical, memristor-based RC in time series prediction,[8,19–25], speech recognition[26], image recognition[27,28] and optoelectronic signal processing[29]. At the same time, it remains unclear how a physical reservoir would deal with input signals containing correlated noise components with correlation times that may be shorter, comparable to, or longer than the reservoir memory range and the forecast horizon.

Here, we test the hypothesis that the performance of a physical reservoir is determined by timescale matching between the three relevant timescales: the noise correlation time, the short-term memory of the reservoir and the forecast horizon of the prediction task.

We investigate a thin oxide film-based physical reservoir with defects that form a solid-state memristor with volatile switching dynamics, as an experimental platform. We studied its performance when dealing with time series injected with synthetic noise and with real-world noisy time series. The synthetic signal is created by mixing a deterministic signal component of two harmonics with added coloured noise. This allows one to independently tune the amplitude and stochastic timescale as a control test. This, in turn, makes it possible to investigate the relationship between the characteristic time of noise correlations of the input data and the fading memory of the physical substrate in physical RC. It is realised by performing the following three tasks:

reconstruction of the deterministic component from a noisy signal, forecasting of a noise-injected signal and forecasting of the coloured noise component alone. We then realise a practical task of forecasting cryptocurrency time-averaged price volatility as a test with a real-world signal which includes both the deterministic and stochastic components.

We then introduce a reservoir memory horizon, which represents the measured timescale of the hardware itself, and a forecasting regime index that describes quantitatively the separation of two forecasting regimes: rapidly varying noise is predominantly averaged by the reservoir, whereas slower correlated noise can be forecasted due to its temporal structure. These results allow establishing new design principles for encoding noisy temporal data for physical hardware and development of novel physical reservoirs.

## Results

### Physical reservoir computing with synthetic noisy signal

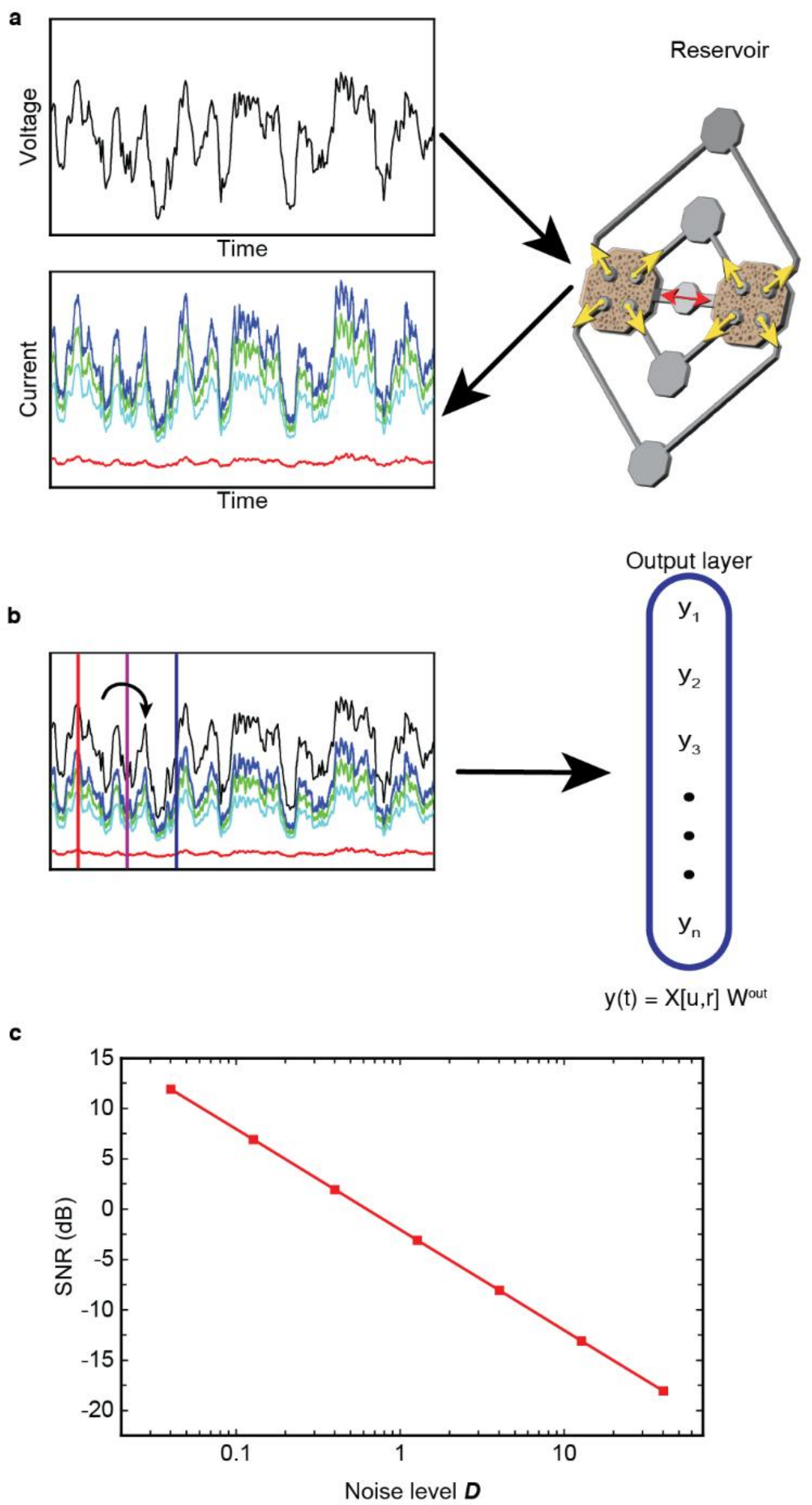

**Fig. 1 | Overview of the reservoir signal processing.** **a**, Voltage signal $u(t)$ is input through the nanoporous niobium oxide physical reservoir by the red arrow and four current outputs $r_j$(t) flow along the yellow arrows. **b**, Voltage $u(t)$ and current $r(t)$ signals are split into past and future windows, example windows are presented by the red, magenta and blue vertical lines splitting the data in the plot. $u(t)$ and $r(t)$ are used, after training the weight matrix $W^{out}$, to predict the future window values of $y(t)$. **c,** Signal-to-noise ratio (SNR) vs the injected coloured noise level $\boldsymbol{D}$ of the deterministic signal.

A nanoporous niobium oxide layer that is sandwiched between two platinum top and bottom electrodes has been shown to act as a capable physical reservoir for temporal time series prediction.[15] A reservoir input was applied to the bottom electrode in the form of a voltage temporal signal, and the reservoir outputs were collected from the top electrodes in the form of electrical current time-dependent signals, which were processed externally via a single perceptron layer trained by linear regression. In this paper, we transitioned to two nanoporous niobium oxide layers which are electrically connected in parallel to enhance the reservoir performance, in comparison to our previously published work[15]. This corresponds to a concept of the parallel reservoir computing[30] and illustrates the flexibility of the in-materia RC approach in terms of testing different RC architectures. The input signal was applied as a voltage signal to two bottom electrodes connected in parallel, and four electrical currents were measured from four pairs of top electrodes at each oxide layer, also connected in parallel, as the reservoir's outputs. That is, two parallel reservoirs (Fig. 1a) with jointly connected nanoporous current channels within the $NbO_x$ thin film layer provided a non-linear conversion of the voltage waveform $u(t)$ to different mutually interconnected current signals $r_j(t)$ which demonstrated short-term, recurring memory. See Fig. S1 for the corresponding current-voltage characteristics of four top electrode channels of the parallel reservoirs.

Fig. 1b demonstrates the RC network operational process whereby a moving window (or time interval) of input data (bottom electrode voltage $\boldsymbol{u}(t)$ and reservoir currents $\boldsymbol{r_j}(t)$ from the top electrodes) is used to predict another window of the target data vector $\boldsymbol{y}(t)$, representing the voltage signal we are interested in. For the training, the matrix $\boldsymbol{X}[u,r]$ is populated by moving the input window across the training range and then used to obtain the matrix of numerical weights $\boldsymbol{W}^{out}$.

Our deterministic, time-dependent voltage function comprises two harmonic terms with the comparable amplitudes $\boldsymbol{V_2 = 0.5V_1}$. The frequencies relationship, $\boldsymbol{f_2 = \sqrt{2}f_1}$ makes its reconstruction and future prediction with reservoir computing non-trivial due to the quasi-periodicity of the signal. Note that we intentionally decided not to use a chaotic signal due to its broad distribution of timescales and partial similarities to noise. Our quasiperiodic deterministic signal has two well-defined time scales which can interfere with correlation time of coloured noise. The total signal was formed by combining the deterministic signal with the coloured noise $\boldsymbol{\eta(t)}$, converting it to a voltage waveform.

$$V_{\text{waveform}} = \boldsymbol{V_1} \cos(2\boldsymbol{\pi} \times \boldsymbol{f} \times \boldsymbol{t}) + \boldsymbol{V_2} \cos\left(2\sqrt{2}\boldsymbol{\pi} \times \boldsymbol{f} \times \boldsymbol{t}\right) + \boldsymbol{\eta}(\boldsymbol{t}) \tag{1}$$

and then fed to the physical reservoir, as depicted in Fig. 1a, through the bottom electrode with four simultaneous currents measurements taken at the reservoir outputs through the top electrodes.

Noise $\boldsymbol{\eta}(\boldsymbol{t})$ in our signal has two parameters: noise correlation time $\boldsymbol{\tau}$, describing how fluctuations decay, and noise intensity $\boldsymbol{D}$ which characterises the strength of fluctuations. This was realised using a standard numerical algorithm described in the Experimental Methods section. The noise correlation times $\boldsymbol{\tau}$ were chosen as $\boldsymbol{\tau} = \boldsymbol{\tau_0}$ to $\mathbf{1000}\boldsymbol{\tau_0}$, $\boldsymbol{\tau_0} = \mathbf{6.3} \times \mathbf{10^{-7}}$s, and the frequency $f$ was chosen to allow ~30 periods during the network training and inference. We also rescale the window sizes with respect to $\boldsymbol{\tau_0}$ to simplify our comparison between different computational tasks outlined below.

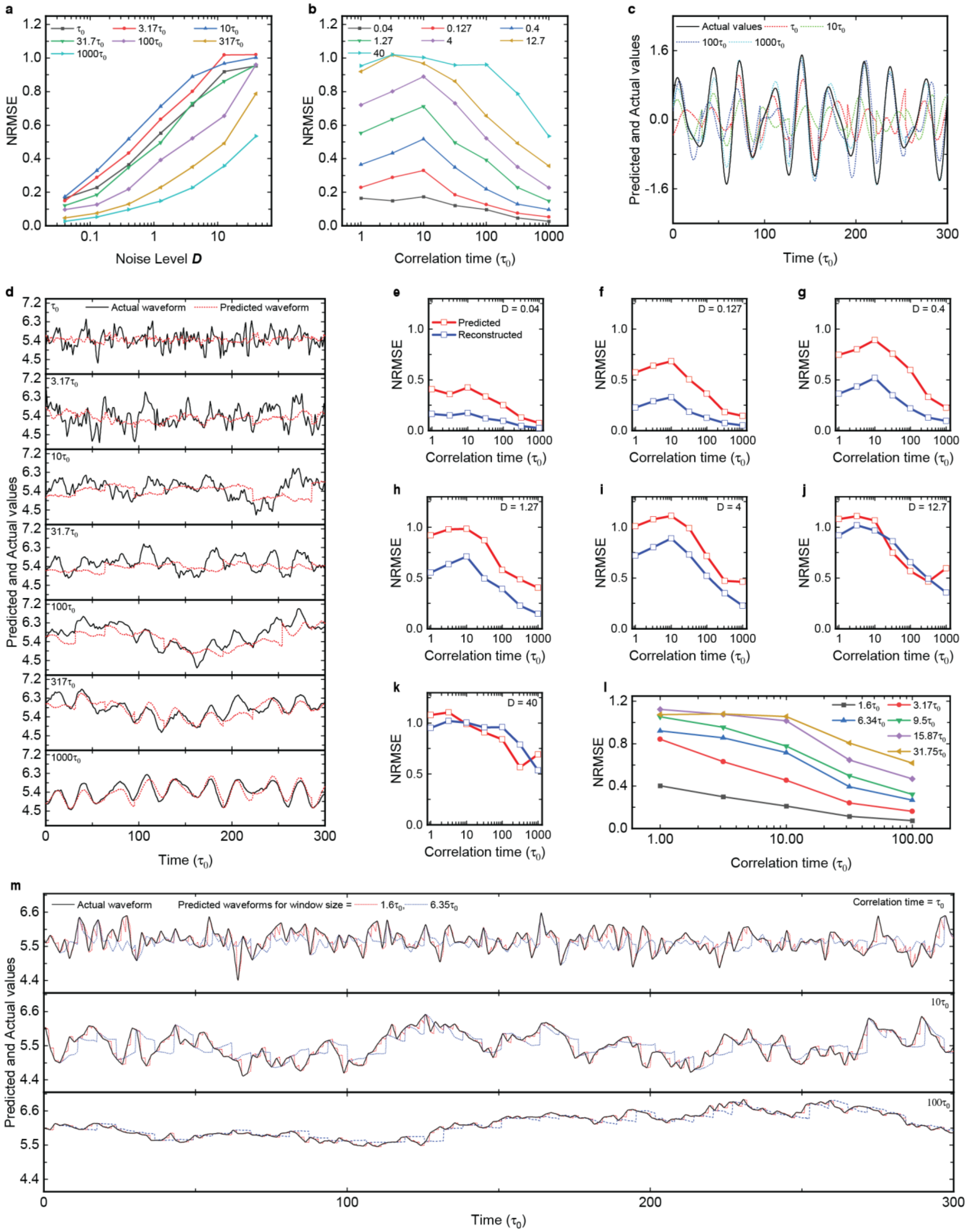

**Fig. 2 | Reconstruction and prediction results for deterministic signals and coloured noise. a,** NRMSE vs noise level ***D*** curves obtained from reconstructing the original deterministic signal for different correlation times $\boldsymbol{\tau} = \boldsymbol{\tau}_0$-$1000\boldsymbol{\tau}_0$, $\boldsymbol{\tau}_0 = 6.3\times10^{-7}$s, for the window size of $31.75\boldsymbol{\tau}_0$. **b,** NRMSE vs correlation time $\boldsymbol{\tau}$ curves obtained for the same reconstruction task as 2**a** for different noise level ***D***=0.04-40. **c,** Comparison of the time-dependent target signal with the reconstructed deterministic signals as obtained from noise injected signals with a noise level ***D***=4 and correlation times $\boldsymbol{\tau}_0$ - $1000\boldsymbol{\tau}_0$, for the window size of $31.75\boldsymbol{\tau}_0$. **d,** Stacked plot of target (solid black line) and future predicted (dotted red line) curves of the coloured noise-injected signal for varying time correlation

$\tau$ and the same noise level $D$=4 from a window size of 31.75$\tau_0$. **e-k,** Comparison of the error NRMSE vs correlation time $\tau$ curves for future predictions of the coloured noise-injected signal (red line) with the reconstructed deterministic signal (blue line) for the same noise level $D$=0.04 (d), 0.127 (e) 0.4 (f), 1.27 (g), 4 (h), 12.7 (i) and 40 (j), respectively. **l,** NMRSE vs correlation time $\tau$ curves for future predictions of the coloured noise alone for various window sizes (past and future window are equal in size). **m,** stacked plot of target (solid black line) and predicted coloured noise curves for the noise level $D$=4 and for window sizes of 1.6$\tau_0$ (dotted red line) and 6.35$\tau_0$ (dotted blue line). The top, middle and bottom plot are for correlation times of $\tau_0$, 10$\tau_0$, and 100 $\tau_0$ respectively.

In order to enable independent comparison with the noise levels represented in this study with the literature sources reported elsewhere[31], we plotted in Fig. 1c the signal to noise ratio (SNR) in terms of average powers (i.e. squares of) of the deterministic signal in relation to the noise term $\eta(t)$ plotted vs noise level $D$ and calculated from $10^6$ time steps. As expected, the SNR is decreasing with increasing $D$ values whilst the noise fluctuations are averaged out.

We first tested the reconstruction ability of the physical RC network, that is, to filter out the noise and to output the original deterministic signal if provided with the deterministic signal injected with coloured noise. The RC network operates with data windows of equal size, 31.75$\tau_0$ for the input (the noise-injected voltage signal) and output data (the deterministic voltage waveform). To quantify the network's performance, the normalised root mean square error (NRMSE) was calculated and plotted against $D$ and $\tau$ parameters in Figs. 2a and 2b, respectively. We found that the error values in Fig. 2a increase with the noise level $D$, as the deterministic signal is obfuscated. At the same time, the smallest error increase with rising noise level $D$ is observed for noise correlation time $\tau$ = 1000$\tau_0$, this meaning the system can best deal with the noise that is governed by the longest correlation times. The largest error increase is observed for $\tau$ close to $10\tau_0$. This is corroborated by Fig. 2b for NRMSE values plotted vs $\tau$ where, for most $D$ values, the NRMSE curves show a broad peak at 10$\tau_0$. A possible explanation for this behaviour is the proximity of the correlation time for the coloured noise to the correlations of the deterministic function with its two period values $1/f$ and $1/\sqrt{2}f$ ($f$ is the basic frequency, $f$=48 kHz) being in the same range as 10$\tau_0$. The idea is that filtering out the noise is at its hardest when it is obscured by deterministic signal varying on a similar timescale. Faster (< 1$\tau_0$) and slower (~1000$\tau_0$) noise dynamics, in comparison to that, are then easier for the system to handle, as can be seen from the direct comparison of the signal curves shown in Fig. 2c (see Fig. S2 and S3 for complete set of reconstructed waveforms). When comparing between these two extreme cases, the slow dynamics yields the lowest NRMSE value. On the other case, the NRMSE curve for $D$ = 0.04 in Fig. 2b demonstrates rather a weak dependence on correlation times $\tau$. So for the weakest noise level case, the noise is rather a small disturbance, which leads to a monotonic decrease of the NRMSE error with increasing correlation time values.

The next test of our physical RC system was to perform prediction of "future" values of the complete waveform function values from the "past" values as the input, see Fig. 1b for the corresponding schematic. All the other parameters are the same as in the section above except the input and output windows are now sequential with respect to each other. Note that the prediction process was a non-autonomous one, so that after each prediction step the input window was moved to the later point in time as part of the sequence, and new data from the input window entered the network and not the previously predicted ones.

Fig. 2d shows a series of predicted waveforms for noise levels $\boldsymbol{D}$ = 4 with different noise correlation time values compared to their target waveform (see Fig. S4 for the plots from other noise levels). Figs. 2(e-k) compare the NRMSE values vs noise correlation time $\boldsymbol{\tau}$ curves for the reconstruction of the deterministic signal and prediction of future values of the noise-injected signal for every noise level $\boldsymbol{D}$. Again, for the predicted signal we see the NRMSE values increase with larger noise levels $\boldsymbol{D}$ and observe the local maxima at the same correlation time of $10\boldsymbol{\tau_0}$ as in the above case of reconstruction. On the other hand, the overall NRMSE values for predictions are higher than NRMSE for reconstruction plotted at the same $\boldsymbol{\tau}$ and $\boldsymbol{D}$ combinations. That can be understood by considering that the network in the case of the prediction is expected not only to filter the noise from the input signal, as for the reconstruction task, but also to infer the future values of the deterministic signal with the injected noise. It is noticeable how the difference in NRMSE errors between the two tasks is the largest for the waveforms with $\boldsymbol{\tau} < 317\boldsymbol{\tau}_0$ and for $D < 1.27$. Therefore, for a relatively small fraction of noise (SNR > 2dB, Fig. 1c), the noise components with higher frequencies (lower $\boldsymbol{\tau}$ values) have the strongest impact on the performance. Contrary to that, for $\boldsymbol{D}$ = 1.27 and $\boldsymbol{D}$ = 4 the prediction performance was worse than the reconstruction one for all $\boldsymbol{\tau}$ values, whilst the highest noise levels $\boldsymbol{D}$ = 12.7 and $\boldsymbol{D}$ = 40 demonstrated the minimal difference. One can say that at higher noise intensities $\boldsymbol{D}$ the prediction performance was generally poor and eventually matched the reconstruction one for the highest $\boldsymbol{D}$ values.

Finally, we studied how well the neural network based on our physical RC system can predict the future values of the coloured noise contribution only, that is, without the deterministic component. Because of its temporal correlations, coloured noise evolves smoothly and persistently in time, providing a structured feature that our system may learn to exploit. The coloured noise was first simulated for the noise amplitude $\boldsymbol{D}$ = 4 with different correlation time values $\boldsymbol{\tau}$ to generate the waveforms and our encoding process rescales them to fit between 4 V and 7 V in the final signal to be applied to the reservoir. However, as a consequence of the encoding process, the absolute value of $\boldsymbol{D}$ had no effect on the noise, as the amplitude was no longer relative to the deterministic function.

As shown in Fig. 2l, the lowest NRMSE values for coloured noise predictions are for the largest correlation time $\boldsymbol{\tau}$. The network performance is again the best for the slowest noise dynamics, in agreement with Figs. 2a - 2k. This is illustrated by the direct comparison of predicted and target curves plotted over time in Fig. 2m (see Fig. S5 for the full set of coloured noise predictions). At the same time, we see an increase in the NRMSE values when the size of the windows are increased (Figs 2l and 2m). For prediction time windows, which are shorter than the noise correlation time, our physical RC has better capability (i.e. lower error values) of predicting the noise compared to when the prediction window size is beyond the noise correlation time. This is particularly visible for correlation times $\geq 10\boldsymbol{\tau}_0$, for example.

**Computing with real-world data**

In the previous parts we studied performance of our RC network using synthetically generated signals where deterministic and noise-related components can be clearly separated by design. This is mostly impossible in the real-world applications. As a test for our system under such conditions, we performed prediction computations for the time-averaged volatility for prices of four cryptocurrencies: Bitcoin (BTC), Ethereum (ETH), Solana (SOL) and Dogecoin (DOGE).

The volatility data set $\boldsymbol{\sigma_n}$ at time step *n* was calculated from the log return functions: $\boldsymbol{R_i} = \mathbf{ln}[\boldsymbol{p_i}/\boldsymbol{p_{i-1}}]$, where $\boldsymbol{p_i}$ signifies the price at the time moment $\boldsymbol{i}$. The volatility requires the averaging of price variations within a specific averaging time interval (averaging window size), $\boldsymbol{\sigma_n} = \sqrt{\langle \boldsymbol{R_i^2} \rangle - \langle \boldsymbol{R_i} \rangle^2}$. By choosing different sizes of the averaging window, we obtained volatility data sets with different correlation times (see figs. S6 and S7 legends for these values), thus linking the volatility time series to the previously considered case of the synthetic deterministic signal injected with coloured noise with a variety of correlation times. However, the correlation time in the cryptocurrency volatility is dependent on the specified averaging window size and cannot be artificially defined.

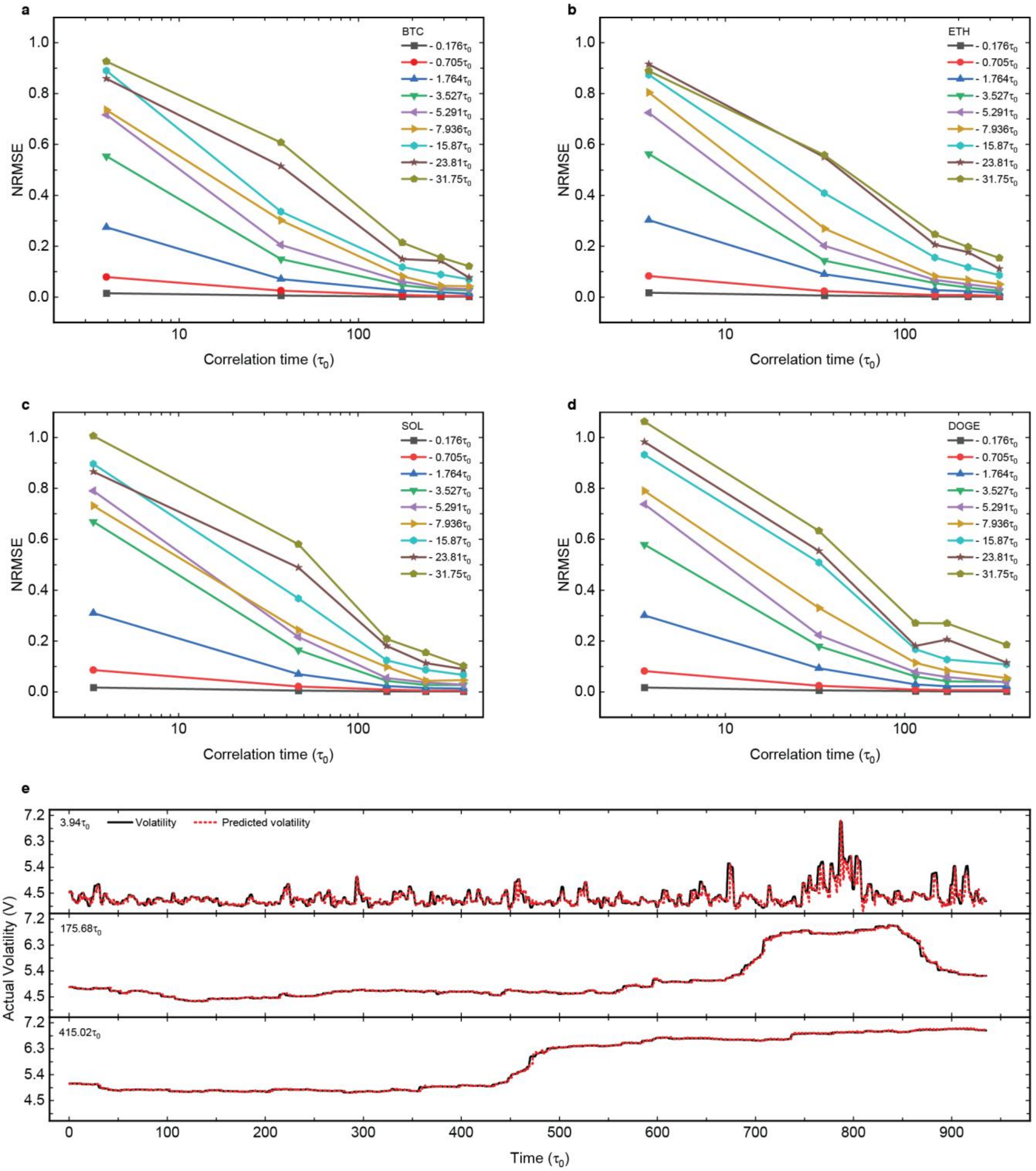

**Fig. 3 | Forecasting of cryptocurrency prices volatility: a-d,** NRMSE vs volatility correlation time $\tau$ curves for four cryptocurrencies, **a**, Bitcoin (BTC), **b**, Ethereum (ETH), **c**, Solana (SOL) and **d**, Dogecoin (DOGE) respectively, for a fixed past window size of $15.87\tau_0$ and future window sizes of $0.176\tau_0$-$31.75\tau_0$. **e,** Comparison of the target (solid black line) and predicted (dotted red line) BTC price volatility vs time for a past window size of $15.87\tau_0$ and a future window size of $3.527\tau_0$, respectively. The top, middle and bottom plots are for a calculated correlation time of $3.94\tau_0$, $175.68\tau_0$ and $415.02\tau_0$ respectively. For the values of the averaging window sizes used to obtain relevant correlation times from the volatility data, see Table 1.

To convert the volatility to a time series applicable to the physical reservoir, 1 h time steps were converted to $1\times10^{-6}$ s and the time axis is plotted further below with respect to the same $\tau_0$. That is, a single $\tau_0$ corresponds to 0.63h in real-world price time series. Table 1 shows the correlation time calculated from the autocorrelation function, for both $\tau_0$ and hourly, for different averaging window sizes.

**TABLE 1.**

| | Correlation time for cryptocurrency price volatilities | | | | | | | |
|---|---|---|---|---|---|---|---|---|
| **Averaging window (hrs / $\tau_0$)** | **BTC (hrs)** | **BTC ($\tau_0$)** | **ETH (hrs)** | **ETH ($\tau_0$)** | **SOL (hrs)** | **SOL ($\tau_0$)** | **DOGE (hrs)** | **DOGE ($\tau_0$)** |
| **3/4.8** | **2.48** | **3.94** | **2.36** | **3.75** | **2.10** | **3.33** | **2.23** | **3.54** |
| **21/33.3** | **23.27** | **36.94** | **22.47** | **35.67** | **29.33** | **46.56** | **20.96** | **33.27** |
| **101/160.3** | **110.68** | **175.68** | **93.03** | **147.67** | **91.16** | **144.70** | **72.38** | **114.89** |
| **201/319.1** | **181.88** | **288.70** | **142.42** | **226.06** | **150.60** | **239.05** | **108.74** | **172.60** |
| **401/636.5** | **261.46** | **415.02** | **212.72** | **337.65** | **244.69** | **388.40** | **233.21** | **370.17** |

The results of forecasting the volatility of cryptocurrencies are shown in Fig. 3 for different correlation times, the same past window size of 15.873$\tau_0$ but different future window sizes. In Figs 3a-d we observe how the NRMSE values decrease as the correlation time increases, in agreement with the previous results on forecasting the synthetic noise-injected signal for longer correlation times. However, no pronounced maxima are observed for the NRMSE curves for the real-world volatility (Fig. 3a-d) in comparison to forecasting and reconstruction of a synthetic signal with a deterministic component (Figs. 2e-k). At the same time, the curves in Fig. 3a-d show strong similarities with the ones for forecasting the coloured noise alone shown in Fig. 2l, see for example the future window size of 15.873$\tau_0$. Similarly, the NRMSE values decrease as the future window sizes decrease. The lowest NRMSE values correspond to the future window sizes smaller than the past window size of 15.873$\tau_0$ but also for correlation times larger than the future window size. This is further illustrated by Fig. 3e which shows comparisons between the predicted (dotted red line) and target (solid black line) signals of the price volatility for BTC for correlation times of 3.93$\tau_0$, 175.69$\tau_0$ and 415.02$\tau_0$ respectively, the same past window size of 15.87$\tau_0$ and the same future window size of 3.527$\tau_0$ (see Figs. S6 and S7 for the similar set of predictions for all four cryptocurrencies).

**Memory capacity and nonlinearity of the reservoir**

In order to evaluate the computational performance of the reservoir, we performed measurements of its memory capacity (MC) and nonlinearity (NL).

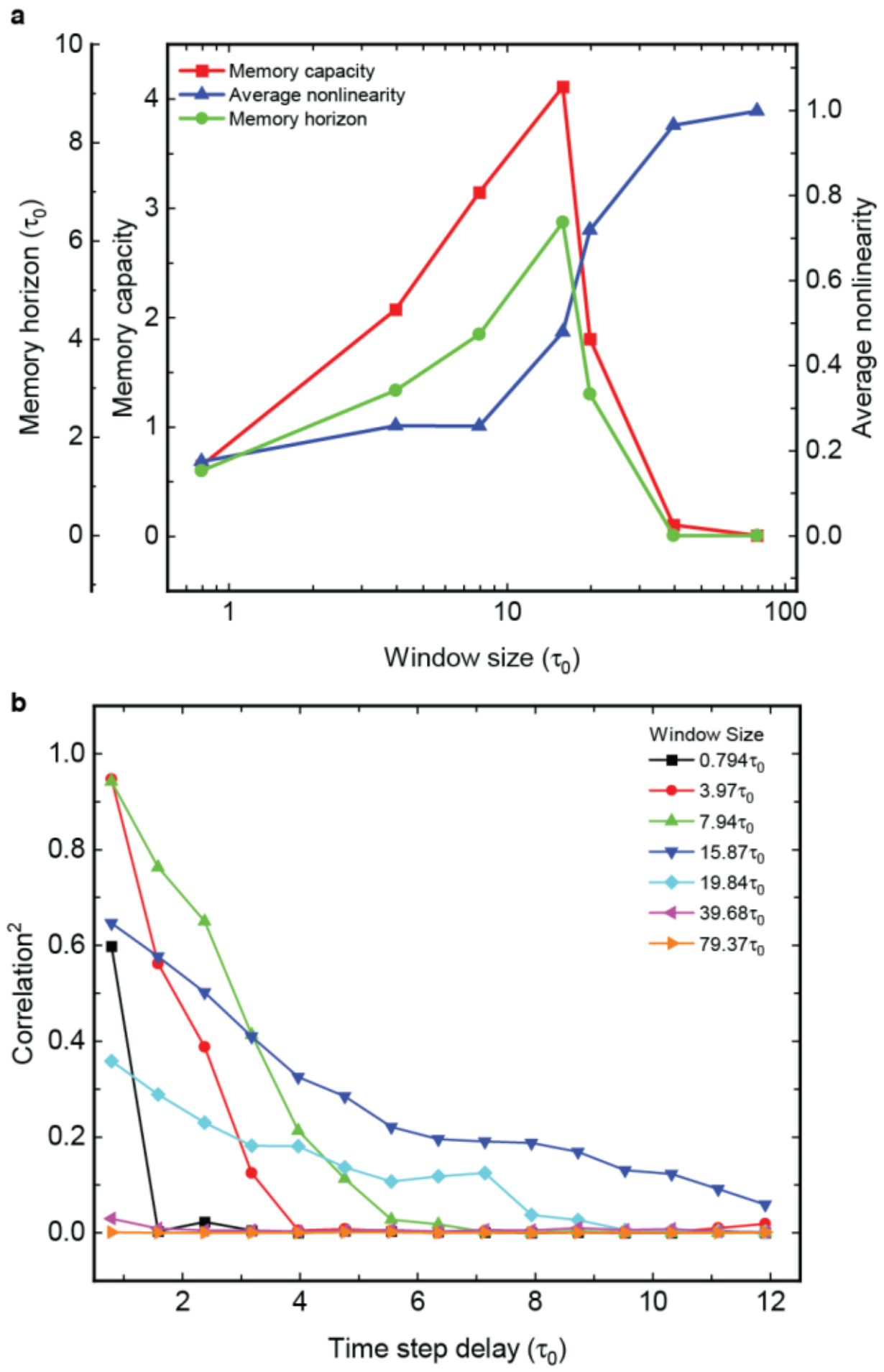

**Fig. 4 | Memory capacity and nonlinearity measurements. a,** Memory capacity (red squares), average nonlinearity (blue triangles) and memory horizon (green circles) vs data window size used in calculating the relevant metrics. **b,** Correlation $\boldsymbol{r^2}$ vs time step delay for varying window sizes.

Fig. 4a shows the effect of the window size that is used in calculating the MC or NL metrics on the MC and NL values. The MC increases from 0.64 to a maximum of 4.11 at a window size of 15.87$\boldsymbol{\tau}_0$ time steps but then decreases sharply for larger window sizes until the largest tested size of 79.37$\boldsymbol{\tau}_0$. A portion of these predicted and target curves are illustrated Fig. S8. Inversely, we see the NL as remaining relatively low up to a window size of 7.94$\boldsymbol{\tau}_0$, after which it increases, getting close to the maximum nonlinearity of 1, for the largest window size of 79.37$\boldsymbol{\tau}_0$. The increase in NL coincides with the post-maximum decrease of the MC.

In order to quantify the time range of the reservoir's short-term memory, we introduce the term of memory horizon (MH) and define it as the time delay range at a given window size (Fig. 4b) for which the corresponding correlation coefficient $\boldsymbol{r^2 \geq 0.2}$. The resulting MH vs window size curve resembles qualitatively the one for the MC, see Fig. 4a. At the same time, by the definition, the linear MC deals with reconstruction of the white noise only and hence, a one-to-one equivalence should not be expected when dealing with the coloured noise instead.

According to Jaeger, the highest achievable MC for a reservoir with linear output functions equals the total number $N$ of internal reservoir units (or neurons) [32]. Further analysis by Jaeger showed that deviations from the reservoir linearity usually result in the reduction from the maximal MC=$N$. Translating this to our system with four reservoir outputs suggests that the outputs (currents) can be better approximated by linear I-V curves when considered at window sizes below or equal to 15.87$\tau_0$ time steps, that is, when the NL remains simultaneously low, and this allows us to demonstrate the MC ≈ 4. Further, when the NL increases at and beyond the processing window size of 15.87$\tau_0$, the MC declines as expected. On the other hand, larger NL is considered necessary when dealing with complex classification tasks, and therefore the window sizes above 15.87$\tau_0$, 19.84$\tau_0$ for example, must have its advantages in the complex tasks despite the relatively small MC. See further discussion in SM (Fig. S9) about a comparison of two window sizes, 31.7$\tau_0$ vs 15.87$\tau_0$ for prediction of noise-injected synthetic signals.

## Two regimes for forecasting noisy signals

Based on the aforementioned results we propose the existence of two regimes for predictions of noisy signals: reservoir-based averaging and algorithm-based prediction.

If the noise correlation times (for example, 1$\tau_0$) are well below both the MH time range (for example, less than 6$\tau_0$) as well as the prediction window size (for example 15.88$\tau_0$), then a reservoir-based averaging is taking place (Fig. 5a, left): the noise component leaves some memory imprint on the reservoir during the network operation but is being averaged out within the prediction window and under the limitations of the reservoir's MH. This means worse overall performance for prediction in comparison to simply reconstructing the deterministic signal, see Figs 2e-k, since the RC network is dealing with the input signal being smoothed out by averaging. The shorter the noise correlation times, the better filtering performance occurs during the same input data window.

If the coloured noise correlation time is longer (for example 317$\tau_0$) than the corresponding MH time (for example 6$\tau_0$) and the prediction window size (for example 15.88$\tau_0$), then the physical reservoir treats the coloured noise component as part of the signal, so that the algorithm is able to reconstruct and predict the signal combined with the injected noise. We call this a regime of algorithms-based forecasting (Fig. 5a, right). The hardware's short-term memory is then well-matched to the input noise correlation times, and the reservoir is able to tolerate the noise and to actively engage with the input dynamics. This explains why the NRMSE values for the reconstruction and prediction as well as the difference between the NRMSE values in the two cases (Figs 2a-k) were at the lowest for the longest noise correlation times.

As a numerical metric to distinguish between the two regimes, we introduce empirically the following forecasting regime index (FRI) which takes into account the ratio of the correlation time of the coloured noise $\boldsymbol{\tau}$ is, to both the MH value $\boldsymbol{\tau_m}$ and the prediction window size $\boldsymbol{\tau_w}$ and is calculated as

$$\text{FRI}=\mathbf{log_{10}}\left[\boldsymbol{\tau}/\sqrt{\boldsymbol{\tau_m \tau_w}}\right]$$

By definition, when the FRI $<$ 0 (FRI>0), the noise evolves faster (slower) than both the prediction and the MH timescale, and the forecasting regime is the one of reservoir-based averaging (algorithm-based forecasting), respectively (Fig, 5b). Further below we illustrate how this metric agrees with the expected forecasting regimes.

In terms of the noise level dependence, the corresponding NRMSE value for prediction of the noise-injected signal at the longest noise correlation time of $\tau$=1000$\tau_0$ (Fig. 5c) has gradually decreased with the decreasing noise level amplitude D towards the ultimate minimum NRMSE=0.00203 for $D$=0, which is for the prediction of the deterministic signal itself.

The two prediction regimes appear in Fig. 2l and Figs 3 a-d where we observe how the coloured noise and the cryptocurrencies volatility price predictions were worse (higher NRMSE values) for the window sizes of 9.5$\boldsymbol{\tau}_0$ and above. But also, the performance was worse at noise correlation times 10 $\tau_0$ and less. Hence the coloured noise with correlation times of less than 15.87$\boldsymbol{\tau_0}$ for the highest MC (longest MH) was filtered out and was predicted relatively poorly within the existing window.

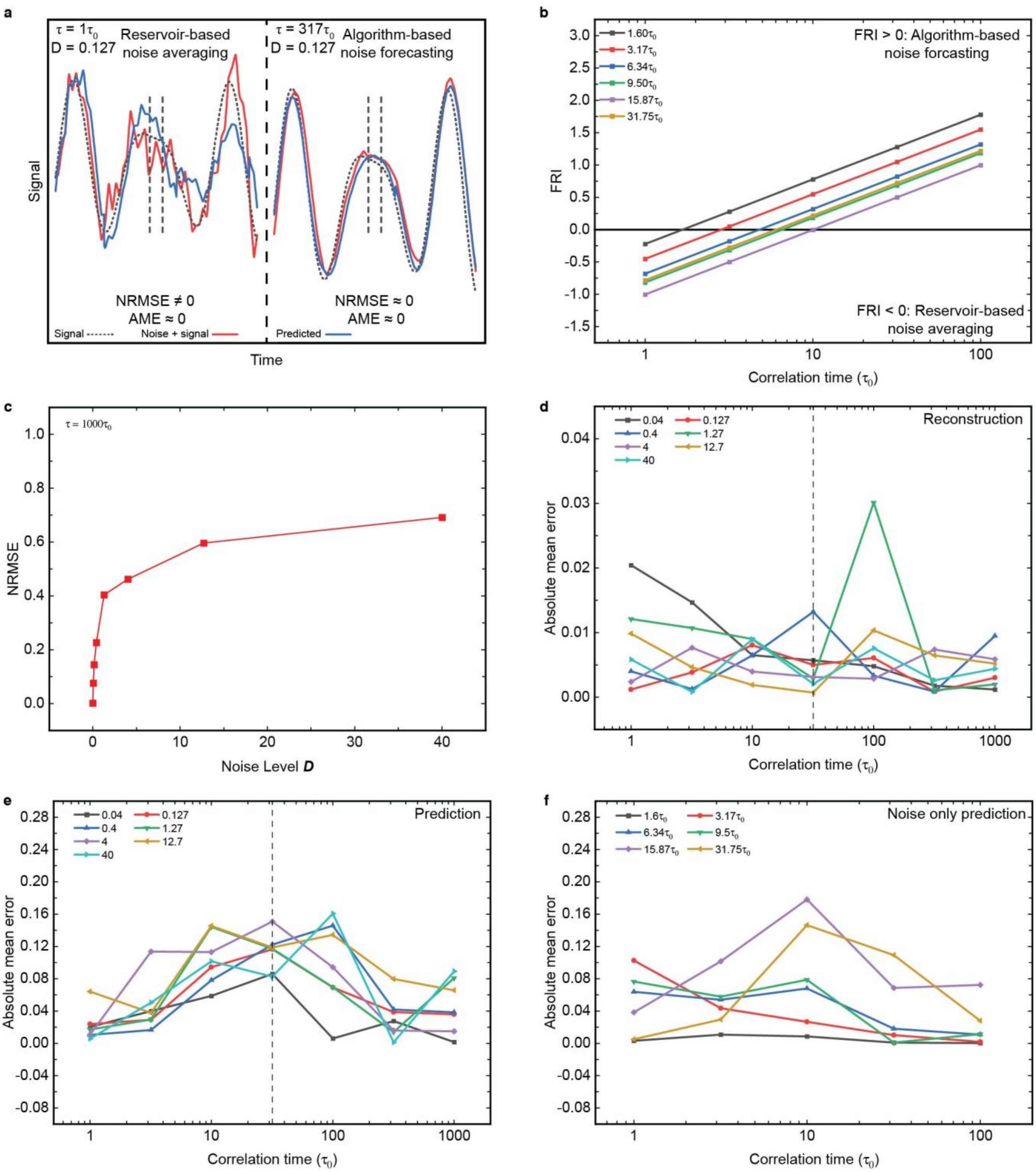

**Fig. 5 | Absolute mean error plots and different noise prediction regimes: a** Illustration of two signal prediction regimes for physical RC, based on the examples of two coloured noise-injected deterministic signals (target, predicted signal and deterministic ones) plotted vs time for noise correlation time $\boldsymbol{\tau} = 1\boldsymbol{\tau}_0$ and $\boldsymbol{\tau} = 317\boldsymbol{\tau}_0$, and a noise level D=0.127. **b** Forecasting regime index (FRI) vs correlation time for different window sizes. **c** NRMSE vs noise level D for the prediction of the coloured noise-injected deterministic signals for a window size = $31.7\boldsymbol{\tau}_0$, and the correlation time $\boldsymbol{\tau} = 1000\boldsymbol{\tau}_0$ taken from Figs 2e-k with an extra point added for the prediction of the deterministic signal itself, D=0. **c-d**, Absolute mean error vs noise correlation time for reconstruction (c), and prediction (d) of noise-injected signals respectively, with the various noise levels D. The dotted vertical line denotes the window size of $31.7\boldsymbol{\tau}_0$ (e) Absolute mean error vs correlation time for predictions of coloured noise only with the different window sizes.

To further illustrate the two regimes for noise-handling by the physical reservoir, we calculated and plotted the additional metric of the absolute mean error (AME), see Figs 5(d-f). The AME is not to be confused with the mean absolute error which is often used as an alternative to NRMSE. The idea behind calculating the AME in this case is that whenever the predicted waveform resembles the deterministic signal which was originally overlapped with the noise in the input signal, then the AME value should approach zero whilst the NRMSE would still return a non-zero value. For example, the AME values for the case of deterministic signal reconstruction (see Fig. 5d, cf. Figs 2a, b) are all fluctuating around zero since in this case the AME is rather a trivial metric, as the target signal is the deterministic one anyway.

The AME values (Fig. 5e) for the prediction of the deterministic signal injected with coloured noise demonstrate a distinct qualitative behaviour: the AME values at the shortest correlation times ($<10\,\tau_0$) are close to zero for noise levels $D < 12.7$, whereas the NRMSE values from the comparable NRMSE vs correlation time curves shown in Figs 2e-k were clearly above zero. Such that the reservoir-based averaging takes place causing a reduction in the AME value.

The AME values in Fig. 5e are again close to zero for the longest noise correlation times, above $100\tau_0$, whereas comparable NRMSE values (Figs. 2e-k) values are at their minimum, too. There, the algorithm-based total signal prediction takes place, since the noise correlation times are longer than both the MH value and the prediction window size ($31.75\tau_0$ in this case, vertical dash line in Figs. 5 d, e).

A qualitatively similar behaviour is found in Fig. 5f for the prediction of the coloured noise-only signal: for the largest prediction window sizes of $15.87\tau_0$ (also corresponds to the largest MC) and $31.75\tau_0$ (also corresponds to high NL) we observe the lowest AME values at the shortest and longest noise correlation time values, respectively, whilst a maximum is visible for the correlation time values in the middle range of the noise correlation times, $10\tau_0$. Note that Fig. 5f is distinctly different from the comparable NRMSE curves shown in Fig. 2l. The maximum in the AME values in Fig. 5f is due to a transition between the two prediction regimes (note the lack of the deterministic signal in the case shown in Fig. 5f, unlike the noise-injected signals shown in Fig. 5d, e), causing the AME to increase. On the other hand, for the prediction window sizes smaller than $15.7\tau_0$, one expects the MH value to be reduced as well, see Fig. 4a, and therefore the peak in Fig. 5f is shifting towards shorter correlation times, so that the reservoir-based averaging regime cannot be observed for those smallest window sizes. Both the occurrence of such transition between the two regimes and its shift at different window sizes agrees qualitatively and quantitatively, to some degree, with the FRI metric we propose and plot in Fig. 5b. There, the zero-crossing of the FRI value indicates the expected transition: for example, the straight line for the window size of $15.7\tau_0$ crosses the timescale axis at $\sim 10\ \tau_0$ (cf. the corresponding peak position in cf. Fig. 5f), and the position of such zero-crossing is shifting to the left with decreasing window sizes.

As a final test, we calculated and compared the correlation times for the coloured-noise predicted waveform (see NRMSE values and AME values in Fig. 2l and Fig. 5f, respectively) with the original noise correlation time value of $1\tau_0$. For the prediction window sizes shorter than $15.87\tau_0$ the predicted waveform yields noise correlation times close to the nominal one (see Fig S10). This is expected since in this window range the MH value of the reservoir is reduced (see Fig. 4a) so no significant filtering should take place. Whereas for the prediction window size of $15.87\tau_0$ and $31.75\tau_0$ we found that the predicted waveforms yielded increased correlation time values $8.26\tau_0$ and $4.89\tau_0$ respectively. That is, in this case the short-correlation time noise is averaged within the prediction window size, and this leads to increase in the noise correlation times obtained from the predicted noise waveform.

## 3 Discussion

The main conceptual advance of this work is that coloured noise is treated not simply as a nuisance, but as a structured component of the learning problem. In many forecasting studies, noise is either weak, uncharacterised, or implicitly assumed to be removable. Here, by independently controlling both the amplitude and correlation time of the coloured-noise contribution, we use noise as a probe of how a physical RC mixed deterministic and stochastic information. This reveals a new view of ML perspectives for physical RC: depending on the relevant timescales, the reservoir can filter noise out or extract predictive structure from it. In this sense, the present memristive system is not only a forecasting substrate, but also a dynamical filter whose behaviour depends on the relationship between input correlations and hardware memory.

A second important result is that the fading memory of the reservoir appears to define a practical operating window for noisy temporal inference. The observed deterioration in performance for larger effective windows and shorter noise correlation times, together with the measured memory capacity, suggests that successful prediction is governed by timescale matching between the input and the hardware. This is especially clear for coloured noise, where the correlation time can be tuned directly. When the noise timescale approaches the relevant dynamical timescale of the deterministic component, prediction and reconstruction become most difficult (Fig. 2b); when the noise evolves more slowly, that is, at time scales longer than the ones for the MH range, then the reservoir can more readily exploit its residual predictability. Whereas at shorter timescales the noise is getting averaged within the prediction window. The cryptocurrency results suggest that this principle remains useful even when deterministic and stochastic components cannot be separated explicitly.

These observations point to a broader design principle for ML hardware: the input signal should be encoded not only to fit the voltage and time range of the hardware, but also to fit its dynamical memory. This is likely to be particularly important when using high-speed physical reservoirs to process slow real-world signals. If the hardware operates at MHz–GHz frequencies while the original data vary on timescales of minutes or hours, then simple uniform rescaling of time may be suboptimal. A more effective strategy may be to compress time non-uniformly, preserving slowly (in comparison to the reservoir's memory timescale) varying segments while shrinking regions with rapid local changes more aggressively. In this view, optimal encoding (ML optimization of encoding process) should be determined by the autocorrelation structure and spectral content of the information signal, spectrum of noise presented in data, and reservoir's memory dynamics. These three factors should determine the encoding process from signal to RC input rather than using the raw sampling interval alone. The goal is to map the informative temporal structure of the signal, including its noise correlations, into the memory timescales where the reservoir is computationally effective.

The reservoir outputs may also contain richer information than is presently extracted by direct regression due to mixture of timescales of input data with time dynamics and filtering of the RC. Different reservoir outputs could in principle be utilised to encode different mixtures of short-term memory, nonlinear transformation and stochastic switching statistics. Future work could therefore treat the output not only as a feature vector for forecasting, but as a structured dynamical representation from which one may decode separate quantities, such as the predicted signal, the reconstructed deterministic component, or an uncertainty estimate associated with the stochastic part.

## 4. Experimental methods

### $\tau_0$ time step comparison

ST1 in the supplementary material shows the comparison between time steps for the different measurements discussed in this work and how they relate to $\tau_0$.

### Physical reservoir:

The nanoporous $N_2$-doped, 50 nm thick niobium oxide layer was sandwiched between the bottom 20 nm thick electrodes of nanoporous platinum and the top electrodes made from titanium adhesion (5 nm) layer and platinum (50 nm) layers which covered an area of 100 $\mu m^2$. The nanoporosity of the platinum bottom electrode introduces heterogeneity to the niobium oxide, aiding the reservoir by increasing the dimensionality and separability. The non-linear *I-V* curve and the volatile resistive and analogue switching behaviour with the corresponding short-term (fading) memory ensures the reservoir is capable of computations that involve time series data, as was previously demonstrated by prediction and reconstruction of chaotic Lorenz-63 time series[15]. The main difference to the previously published reservoir was the increased complexity: each input voltage waveform was fed simultaneously and in parallel into two oxide reservoirs.

**Coloured noise signal:**

$$\boldsymbol{V(t) = V_1 \cos(2\pi \times f \times t) + V_2 \cos\left(2\sqrt{2}\pi \times f \times t\right) + \eta(t)}$$

The basic frequency *f* = 48 kHz was chosen to generate a 2ms long pulse with approximately 2000-times steps. The signal was further injected with coloured noise $\boldsymbol{\eta(t)}$ (in volts) that demonstrates an exponential autocorrelation function $\boldsymbol{\langle \eta(t_1)\eta(t_2)\rangle = D \cdot Var(\xi)\, \exp(-|t_1 - t_2|/\tau)}$ where $\boldsymbol{\tau}$ is noise correlation time in seconds and $\boldsymbol{D}$ is the dimensionless noise intensity. $\boldsymbol{\eta(t)}$ was calculated via solving the Ornstein-Uhlenbeck process by using the Euler-Maruyama method where $\boldsymbol{\eta_0}$= 0V for the starting solution, time step $\boldsymbol{\Delta t}$ and the (*n*+1)-th step of the noise function:

$$\boldsymbol{\eta_{n+1} = e^{-\frac{\Delta t}{\tau}} \times \eta_n + \sqrt{D\left(1 - e^{\left(-\frac{2\Delta t}{\tau}\right)}\right)}\, \xi_n}$$

It guarantees the statistical variance $\boldsymbol{Var(\eta) = D \cdot Var(\xi)}$ for infinitely large samples. $\boldsymbol{\xi_n}$is the *n*-th random voltage value where $\boldsymbol{\xi_n}$ has a Gaussian probability density with a mean of 0V and a standard deviation of 1V. Our results deal with functions *V*(*t*) built using different combinations of values for the noise amplitude *D* and the time correlation $\boldsymbol{\tau}$: $\boldsymbol{\tau = 1\tau_0}$ to $\boldsymbol{1000\tau_0}$. $\boldsymbol{\tau_0 = 6.3 \times 10^{-7}}$s and $\boldsymbol{D = 0.04}$ to $\boldsymbol{40}$. In addition, the total waveform duration and the deterministic signal frequency *f* were chosen to represent ~1000×$\tau_0$ and to allow ~30 periods of *1/f* during both the network training and the inference.

Once the noise and deterministic function were calculated, the resulting waveform was encoded by scaling between 4 V and 7 V to better fit into each device's resistive switching *I-V* curve (see Fig. S1), that means the nominal value of noise amplitude $\boldsymbol{D}$ must be considered in relationship to the deterministic signal and not taken as an absolute.

**Device fabrication and measurement**

Device fabrication for the nanoporous platinum did not change from our previous method except for the lithography pattern used[15]. The $N_2$-doped $NbO_x$ was deposited by magnetron sputtering and patterned by UV lithography. An $Ar/O_2/N_2$ mixture with flow rates of 4 sccm, 9.3 sccm and 0.6 sccm respectively and a working pressure of 2.5-3.6 mTorr were used to deposit the oxide layer. The bottom electrode area under each nitrogen-doped niobium oxide layer has an area of 1325 $\boldsymbol{\mu m^2}$. The $\mathbf{N_2}$-doped niobium oxide layer that covers the bottom electrode has an area of 2982 $\boldsymbol{\mu m^2}$ and a thickness of 50 nm.

These measurements were conducted using a Keithley 4200ASCS by applying the input signal through the common bottom electrode and taking an electric current measurement out of the four top electrode pads simultaneously. An 8 ms long, 5 V high voltage pulse was applied before the modified waveform to pre-warm the reservoir.

### Linear Regression

Linear regression was carried out in the same manner as described in our previous work[15]. To briefly summarise the process, the four current outputs and input waveform (if relevant) were taken from the physical reservoir and split into training and testing data. Bayesian optimization was utilised to help minimise the NRMSE by varying the regularisation term $\boldsymbol{\lambda}$ between $\mathbf{10^{-5}}$ and 100 for synthetic signals or between $\mathbf{10^{-5}}$ and 10000 for the volatility predictions. The number of iterations varied between experiments; each iteration uses a new $\boldsymbol{\lambda}$ in an attempt to minimise the NRMSE.

### Reconstruction and prediction of noise-injected and pure coloured noise signals

The discrete time steps recorded by the Keithley 4200A-SCS were converted with respect to $\tau_0$ = 630 ns. This conversion allows for easier comparison between the different time series measured. The Keithley 4200A-SCS measured a total of 18423 points corresponding to the time duration of 3249$\boldsymbol{\tau_0}$. In order to compensate for inconsistencies in the sampling rate of Keithley 4200A-SCS, the measured waveforms were resampled using the MATLAB function "resample" with a desired sampling frequency determined by the inverse of the average time step. This process was conducted for every measurement. The first 223 points (39.33$\boldsymbol{\tau_0}$) were cut-off as a wash out period leaving a total of 10800 time steps (1905$\boldsymbol{\tau_0}$) for training, which is 58.6% of the data. The test window starts 1306 steps (230.33$\boldsymbol{\tau_0}$) after the training window ends, in order to keep the windows temporally separate. This leaves a final data test size of 5400 (952.36$\boldsymbol{\tau_0}$) which is 29.3% of the total data. This gives a 2:1 ratio of training to testing data ranges. The moving window for the reconstruction and prediction has a size of 180 time steps (31.75$\boldsymbol{\tau_0}$). The washout period, the temporal separation and some data points cut-off at the end account for the remaining 12% of the total data, which went unused. The maximum number of iterations for Bayesian optimization was set to 200.

### Memory capacity and non-linearity of the reservoir

The fading memory capability of the reservoir is characterised by the so-called linear MC:

$$\boldsymbol{MC = \sum_{k=1}^{\infty} r^2\,[u(n-k), y_k(n)]}$$

which is calculated a sum of the squares of the correlation between predicted and original signal for all time step delays *k*, where a white noise signal $\boldsymbol{u(n)}$ is taken as the input and sent through the reservoir, then delayed by *k* time steps to form a new, delayed target signal $\boldsymbol{u(n-k)}$ which is then used to train the network with the reservoir outputs produced by $\boldsymbol{u(n)}$ and to generate the corresponding network output $\boldsymbol{y_k(n)}$.

Nonlinearity (NL) of the reservoir inversely, is calculated using the following formula[33].

$$\boldsymbol{\Phi_n^{NL} = 1 - r^2\left[\Sigma_{ij} W_{ij} u(n), \hat{y}(n)\right]}$$

where the reservoir's current $\boldsymbol{\hat{y}(n)}$ is re-created as a linear combination $\boldsymbol{\Sigma_{ij} W_{ij} u(n)}$ using the reservoir's input voltage $\boldsymbol{u(n)}$ with numerical weights $\boldsymbol{W_{ij}}$. A reduction in the corresponding correlation $\boldsymbol{r}^2$ coefficient implies an existing non-linearity of the reservoir's current with respect to the reservoir's input voltage.

As an input signal, a 1.02 ms Gaussian white noise waveform was generated with the time between each step 0.5 $\boldsymbol{\mu}$s. As in the previously discussed task, the waveform was encoded by rescaling between the bounds of 4 and 7 V to be compatible with the reservoir.

The measured waveforms were resampled from 18423 data points down to 2047 data points (both 1624.6$\boldsymbol{\tau_0}$) using the built-in MATLAB R2024a function "resample", in order to match the number of originally defined white noise values generated for the input waveform.

The delay, $\boldsymbol{k}$ , was tested from 0.794$\boldsymbol{\tau_0}$ to 11.905$\boldsymbol{\tau_0}$ time steps for a range of windows size of 1, 5, 10, 20, 25, 50 and 100 time steps (0.794$\boldsymbol{\tau_0}$, 3.968$\boldsymbol{\tau_0}$, 7.936$\boldsymbol{\tau_0}$, 15.873$\boldsymbol{\tau_0}$, 19.841$\boldsymbol{\tau_0}$, 39.682$\boldsymbol{\tau_0}$, 79.365$\boldsymbol{\tau_0}$). A total of 300 data points (238.09$\boldsymbol{\tau_0}$) were used for training and 100 (79.366$\boldsymbol{\tau_0}$) for testing. The training data consisted of 300 time steps (238.10$\boldsymbol{\tau_0}$) starting from time step 372 (295.23$\boldsymbol{\tau_0}$) to allow for a washout period. The test range was 100 time steps (79.37$\boldsymbol{\tau_0}$) long starting from time step 1486 (1179.4$\boldsymbol{\tau_0}$). This ensured there was temporal separation between the training and testing windows.

The maximum number of iterations for Bayesian optimization was set to 300.

**Cryptocurrency volatility**

Resampling was done in the same way as the noise reconstruction and prediction step. Training and test window sizes were also defined the same way. The maximum number of iterations for Bayesian optimization was set to 100. The price was taken every hour from the date of 11.05.2024 for 85.3 days in total. The full range was encoded into a waveform usable by the Keithley-4200A-SCS by scaling between 4 and 7 V. Volatility was calculated using in MATLAB R2024a. For more details see the supplementary text.

Supplemental material

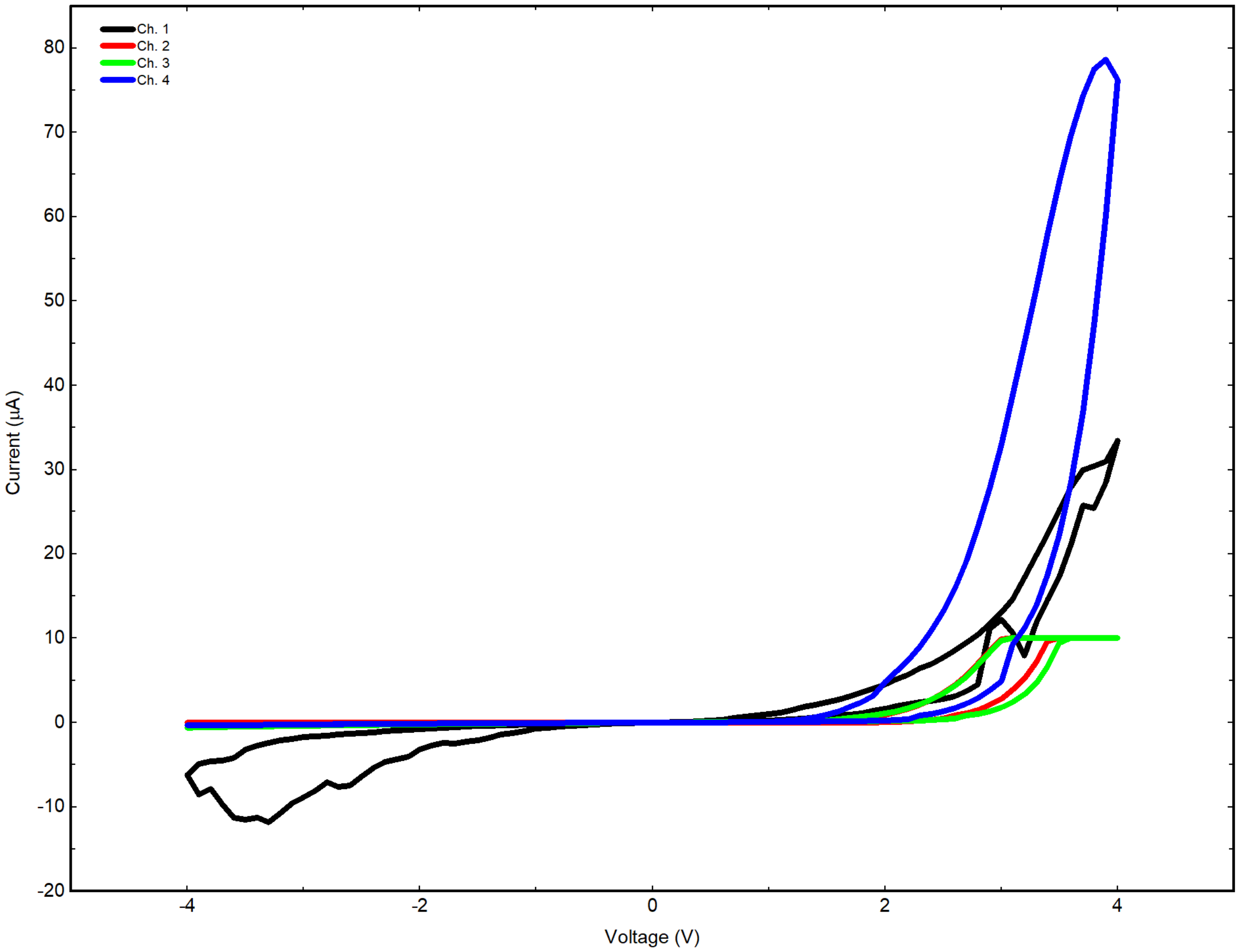

Fig. S1 | **Current-Voltage behaviour of nanoporous oxide memristors:** Current-voltage characteristic curves from the four pairs of top electrodes of two parallel reservoirs. The voltage was swept from 0 to 4 V, then to –4 V and back to 0 V in steps of 0.1 V. Channel 2 and 3 limited by compliance current.

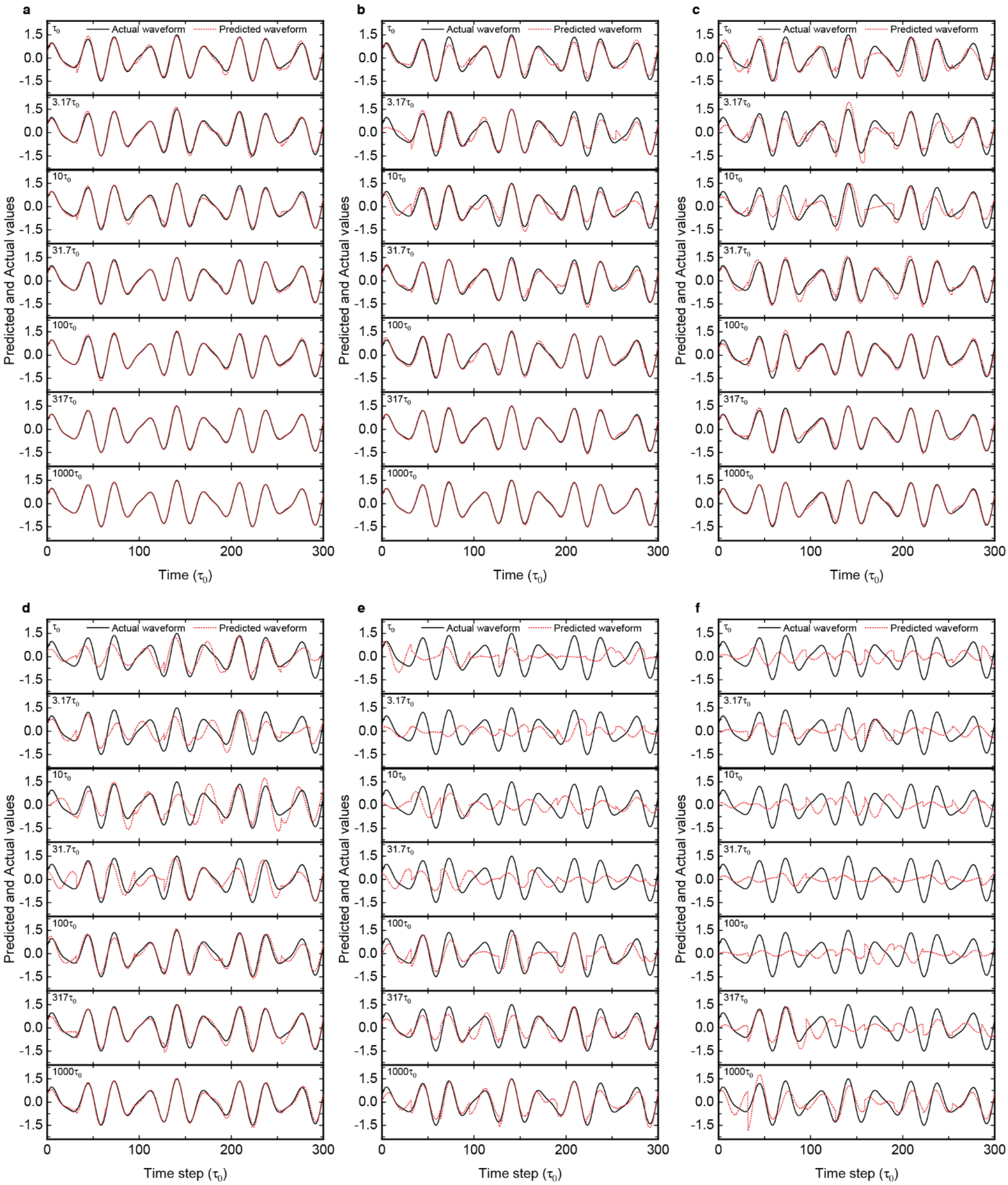

Fig. S2 | **Reconstruction of deterministic signal vs time. a-f,** reconstruction from different noise levels $\boldsymbol{D}$ = 0.04 (a), 0.127 (b), 0.4 (c), 1.27 (d), 4 (e), 12.7 (f) respectively. The first prediction in each subplot is for the smallest coloured noise correlation time $\boldsymbol{\tau_0}$, increasing up to $1000\boldsymbol{\tau_0}$ down each subplot. The target signal and reconstructed waveform are denoted by the solid black line and dotted red line respectively.

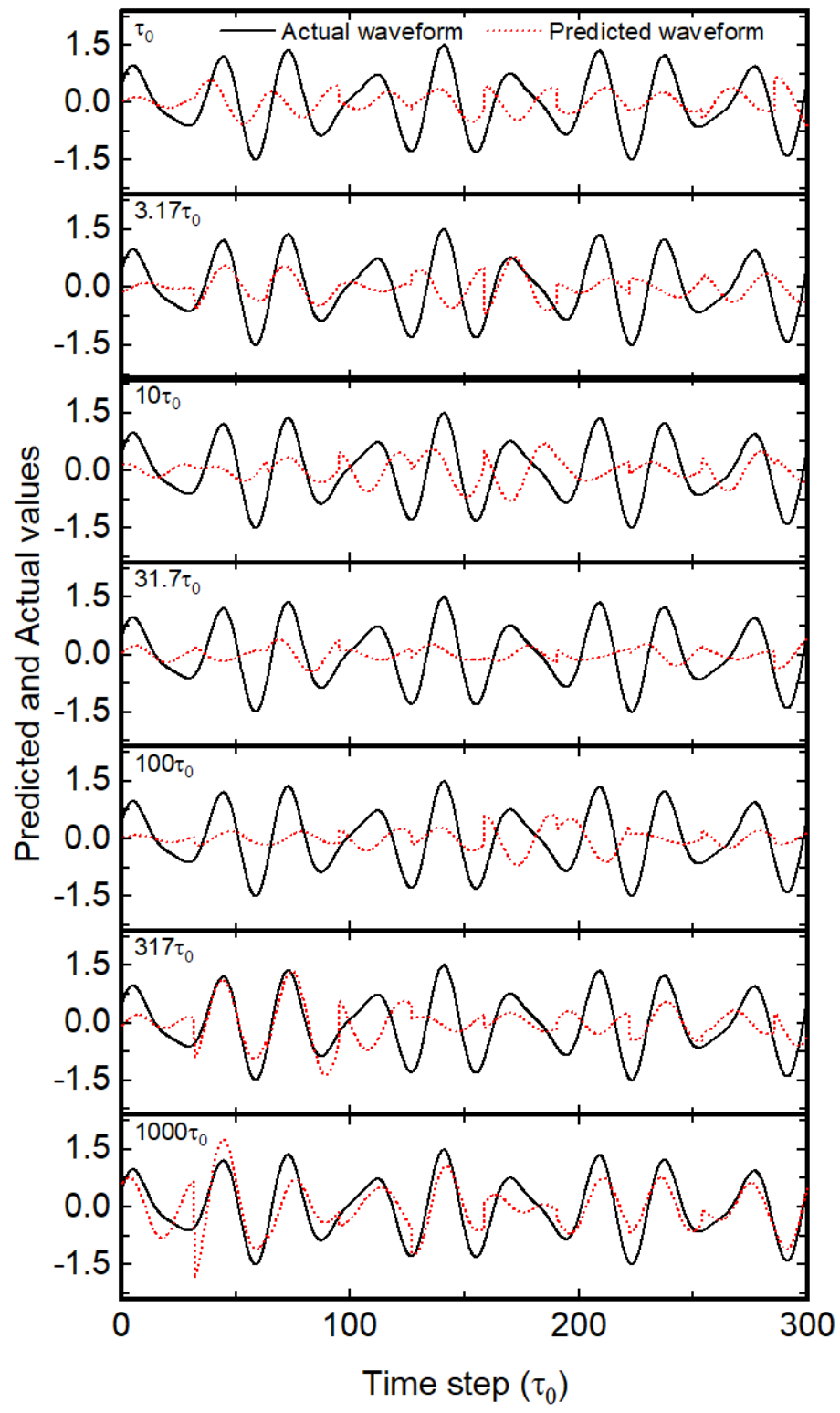

Fig. S3 | **Reconstruction of deterministic signal vs time.** The same type of curves as the ones shown in Fig. S2 plotted for a noise level $\boldsymbol{D} = 40$.

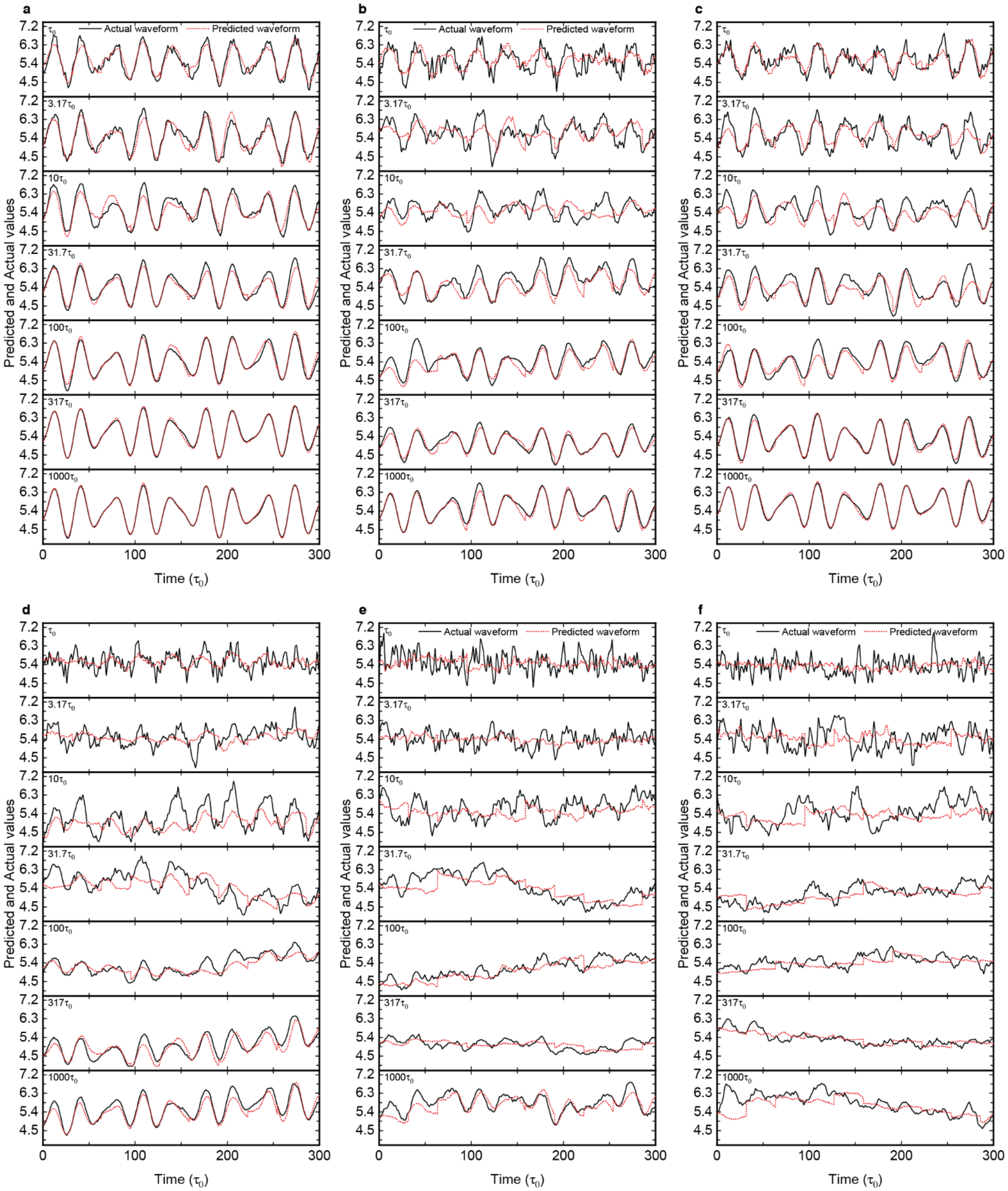

Fig. S4 | **Prediction of the noise-injected signal vs time. a-f,** predictions from different noise levels $\boldsymbol{D}$ = 0.04 (a), 0.127 (b), 0.4 (c), 1.27 (d), 12.7 (e) and 40 (f) respectively ($\boldsymbol{D}$ = 4 has been omitted here as it is already shown in Fig. 2d). The first prediction in each subplot is for the smallest coloured noise correlation time $\boldsymbol{\tau_0}$, increasing up to 1000$\boldsymbol{\tau_0}$ down each subplot. The target signal and predicted waveform are denoted by the solid black line and dotted red line respectively.

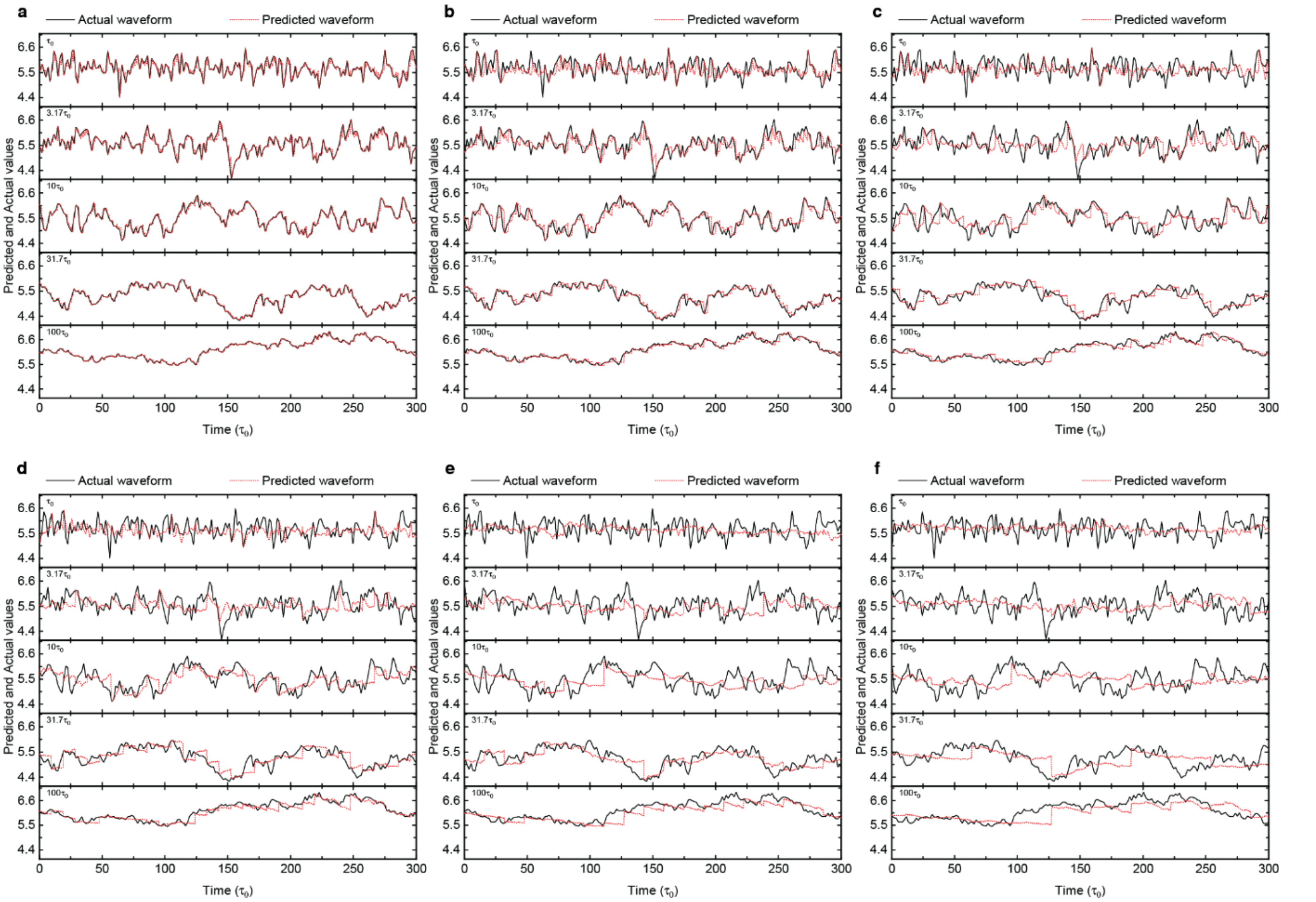

Fig. S5 | **Prediction of coloured noise signal vs time. a-f,** predictions from different window sizes = 1.6$\tau_0$ (a), 3.17$\tau_0$ (b), 6.34$\tau_0$ (c), 9.5$\tau_0$ (d), 15.87$\tau_0$ (e), 31.75$\tau_0$ (f) respectively. The first prediction in each subplot is for the smallest coloured noise correlation time $\boldsymbol{\tau_0}$, increasing up to 1000$\boldsymbol{\tau_0}$ down each subplot. The target signal and predicted waveform are denoted by the solid black line and dotted red line respectively.

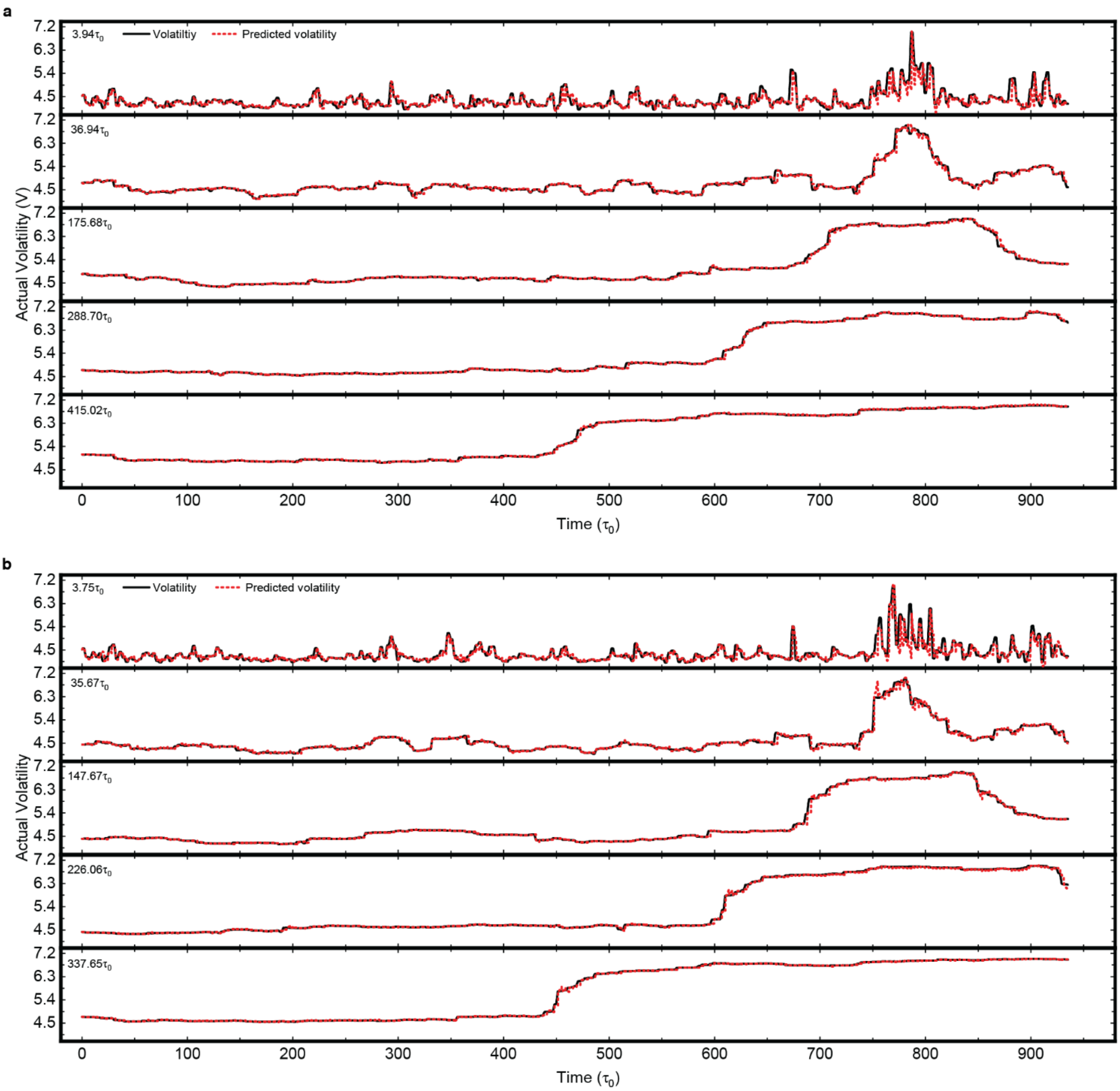

Fig. S6 | **Forecasting of cryptocurrency prices volatility.** Comparison of the target (solid black line) and predicted (dotted red line) price volatility for Bitcoin (a) and Ethereum (b) vs time for a past window size of $15.87\tau_0$ and a future window size of $3.527\tau_0$, respectively, for various correlation times. For both plots, each subplot is for a volatility produced by an averaging window sizes of $4.762\tau_0$, $33.333\tau_0$, $160.317\tau_0$, $319.05\tau_0$ and $636.51\tau_0$ respectively. The value in the top left of each subplot shows the calculated correlation time. Note how it varies between the first subplot in **a** and **b**, so correlation time is not one to one related to the size of the averaging window.

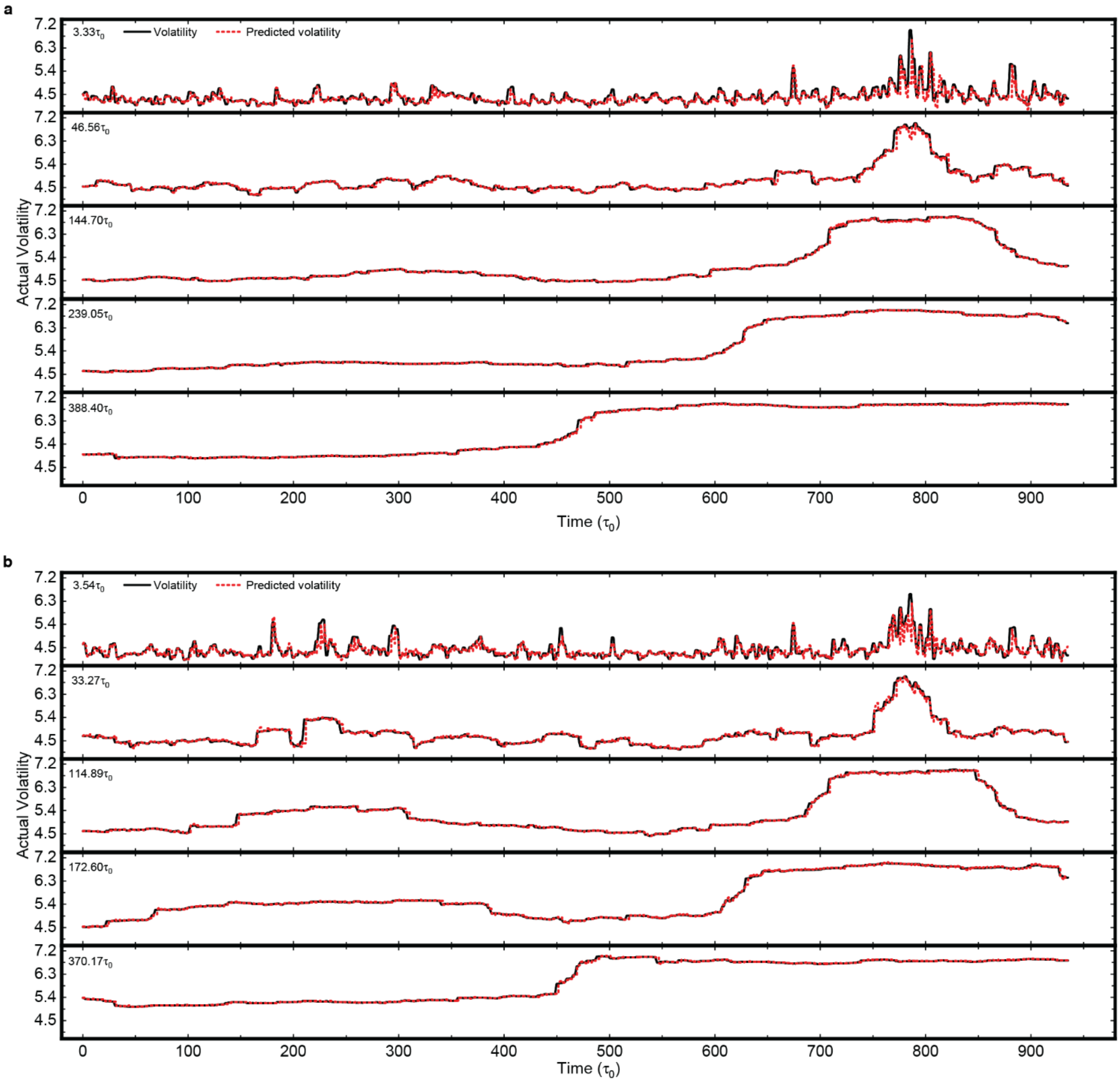

Fig. S7 | **Predicted volatility of cryptocurrencies.** Similar plots as Fig. S6 for the prediction of **a** Solana and **b** Dogecoin. The target (solid black line) and predicted (dotted red line) price volatility.

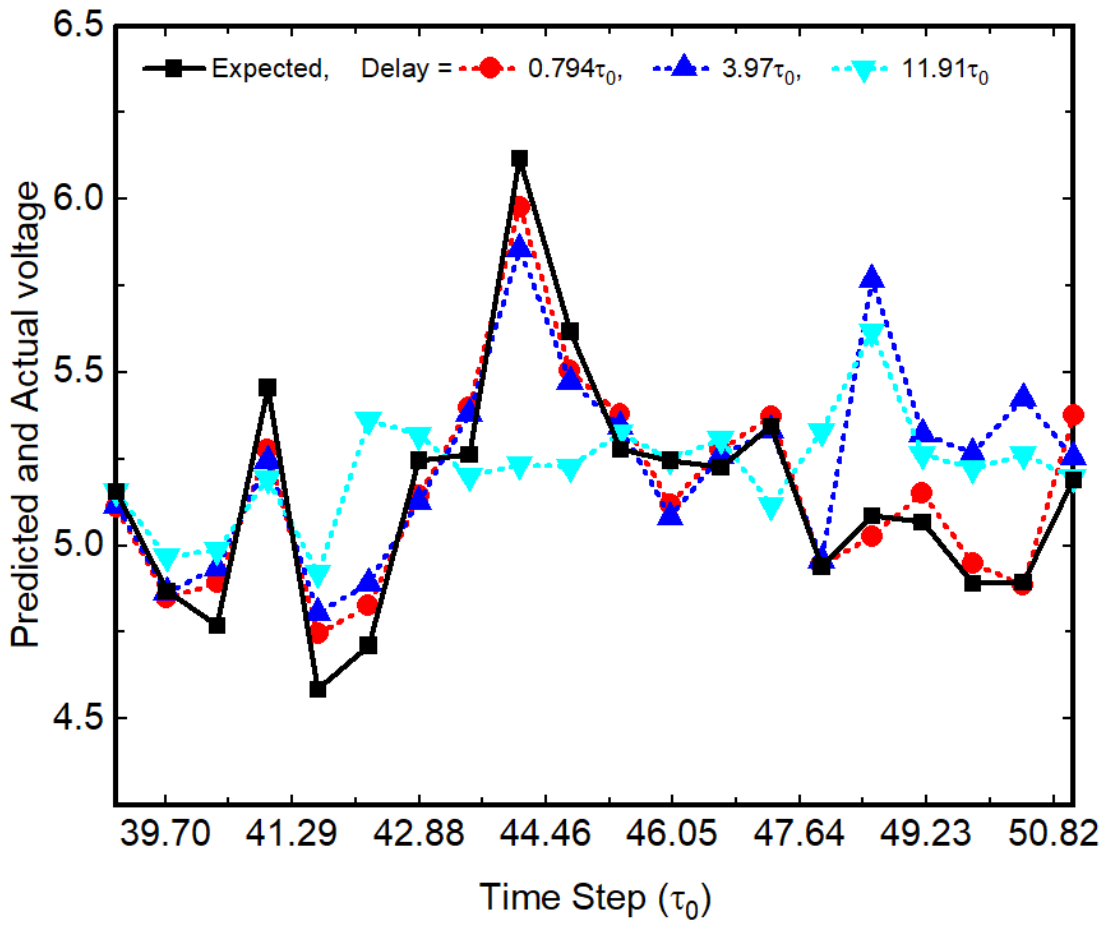

Fig. S8 | Reconstruction of delayed **white noise waveform:** Comparison between the predicted and the target white noise input signal (black squares) vs time step for a window size of 15.87$\tau_0$ and delay values of 0.794$\tau_0$, 3.968$\tau_0$ and 11.905$\tau_0$ respectively. Predicted waveforms are shifted to the same time steps range as for the delay $k = 0.794\tau_0$.

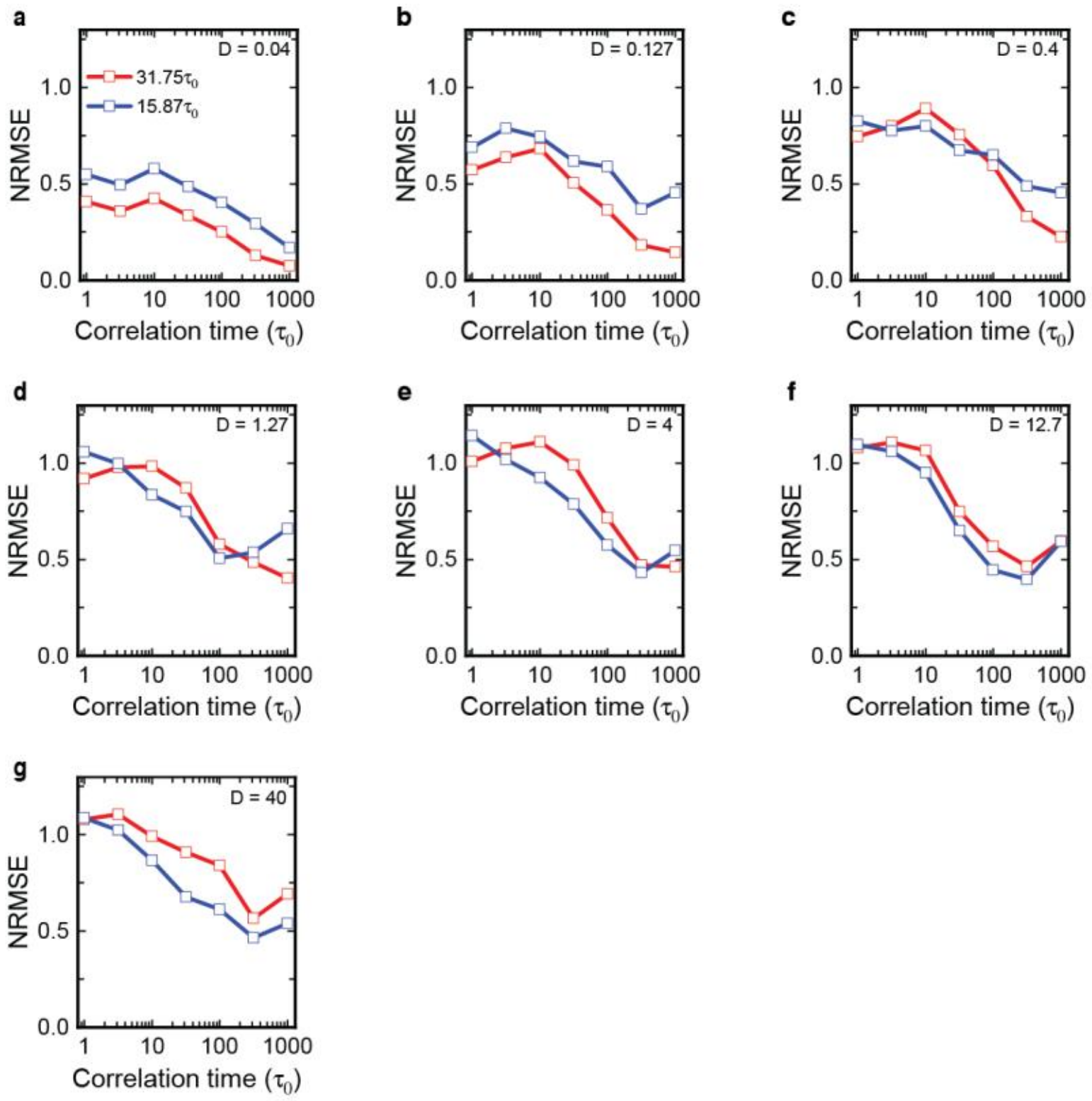

Fig S9. **| Prediction of the noise-injected signal vs time:** NRMSE vs correlation time for two window sizes, $31.75\tau_0$ (red line) and $15.87\ \tau_0$ (blue line) and varied noise level D, 0.04 (a), 0.127 (b), 0.4 (c), 1.27 (d), 4 (e), 12.7 (f) and 40 (g) respectively.

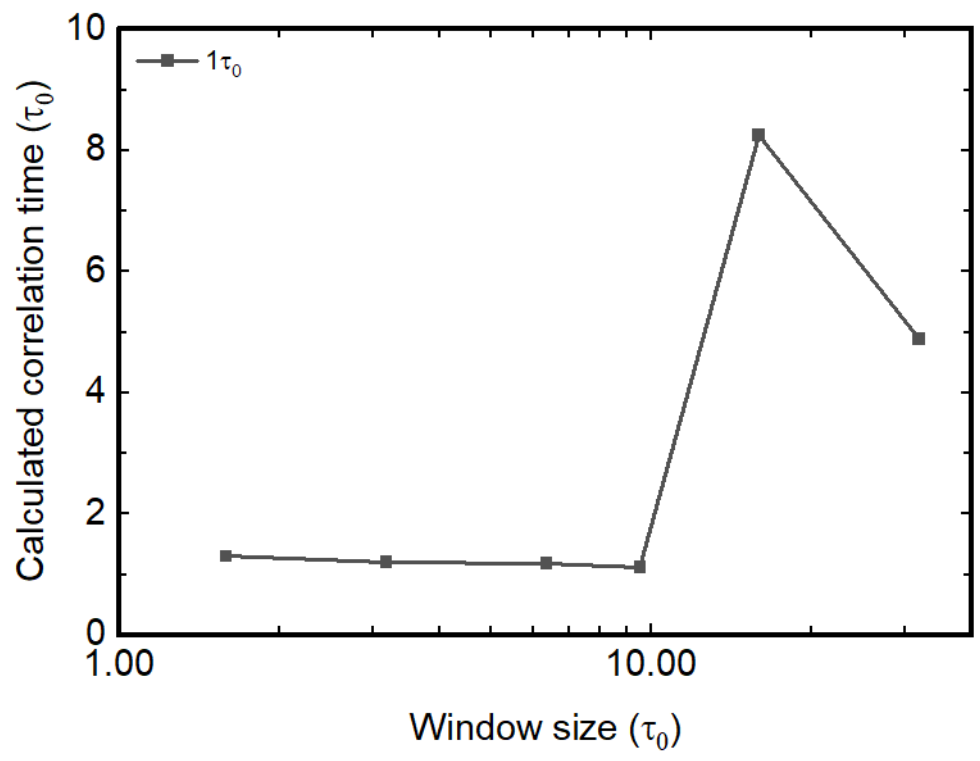

Fig. S10 | **Noise correlation time vs window size:** Noise correlation time obtained from the coloured noise waveforms which were predicted from a waveform with a nominal noise correlation time of $1\tau_0$, vs the prediction window size.

| | MC | Noise | Volatility |
|---|---|---|---|
| Average time step (μs) | 0.5 | 0.111 | 0.111 |
| Time step in $\boldsymbol{\tau}_0$ | 0.794 | 0.176 | 0.176 |

ST1. Average time steps for the individual measurements in seconds and with respect to $\boldsymbol{\tau}_0 = \mathbf{6.3 \times 10^{-7}}$s.

Supplementary Text

Volatility calculations

Log returns $\boldsymbol{R_i}$ at the time point $\boldsymbol{t_i}$ are calculated as

$$\boldsymbol{R_i} = \ln\left(\frac{\boldsymbol{p_{ti}}}{\boldsymbol{p_{i-1}}}\right)$$

The demeaned log returns data are then processed using averaging windows of size $\boldsymbol{t_{aws}}$

$$\boldsymbol{t_n} - \frac{\boldsymbol{t_{aws}}}{\boldsymbol{2}} \leq \boldsymbol{t_n} < \boldsymbol{t} + \frac{\boldsymbol{t_{aws}}}{\boldsymbol{2}}$$

Where $\boldsymbol{t_n}$ is the time step for the corresponding the volatility value, assigned to the middle $\boldsymbol{t_{aws}}$. Volatility $\boldsymbol{\sigma(n)}$ is calculated as:

$$\boldsymbol{\sigma_n} = \sqrt{\langle \boldsymbol{R_i^2} \rangle - \langle \boldsymbol{R_i} \rangle^2}$$

The volatility $\boldsymbol{\sigma}$ is also demeaned to obtain $\boldsymbol{\widetilde{\sigma}}$.

A number of MATLAB functions were employed for finding the autocorrelation time: the autocorrelation function (ACF) was calculated as real part the inverse Fourier transform (FFT),

$$\boldsymbol{ACF = real\big(ifft(PSD)\big)},$$

of the power spectral density (PSD), which in turn was calculated through Fourier transforms as the following:

$$\boldsymbol{PSD = fft|FFT(\widetilde{\sigma})|.^2}$$

$$\boldsymbol{FFT(\widetilde{\sigma}) = fft(\widetilde{\sigma}, nFFT)}$$

$$\boldsymbol{nFFT = 2^{\left(nextpow2\left(length(\widetilde{\sigma})\right)+1\right)}}$$

where nFFT is the zero-padded FFT length which helps create a smoother frequency spectrum. The MATLAB function "$\boldsymbol{nextpow2}$" finds the smallest power of 2 that is greater than the length of the calculated volatility $\boldsymbol{\widetilde{\sigma}}$ + 1.

In order to obtain an estimate for the autocorrelation time, we normalised the ACF by its first value

$$\boldsymbol{ACF_{normalised}} = \frac{\boldsymbol{ACF}}{\boldsymbol{ACF(1)}}$$

and then applied a linear fit up to the cutoff correlation at a threshold of $\frac{1}{e}$ which holds the purpose of stop using correlations which are indistinguishable from noise. The correlation time was then calculated as the negative inverse of the fitted slope.

**Comparison of two window sizes for reconstruction and prediction of coloured noise-injected signals**

The noise-injected deterministic signal is predicted and reconstructed within the window size of $31.7\boldsymbol{\tau_0}$ (Fig. 2a-k) which has a reduced MH but also corresponds to an increased NL in the reservoir response (Fig. 4a), so that the deterministic signal was reconstructed to an acceptable degree of error. We also ran tests on reconstruction and prediction of coloured noise-injected signals for the window size of $15.87\boldsymbol{\tau_0}$ corresponding to the largest MH (see Fig. S9) and found no remarkable difference in the network performance for both cases. In the comparison of the two cases, lower NRMSE values are observed for the window size of $31.7\boldsymbol{\tau_0}$ for the noise amplitude values $D$<1.27 (SNR>-3 dB, Fig. 1c), whilst lower NRMSE values were seen for larger noise amplitudes $D$>1.27 (SNR<-3 dB, Fig. 1c) in the case of window size of $15.87\boldsymbol{\tau_0}$ and longer correlation times $\boldsymbol{\tau}$**>** $10\boldsymbol{\tau_0}$**.** That is, a moderate MC and higher NL in the case of the window size of $31.7\boldsymbol{\tau_0}$ had an advantage over slightly larger MC and lower NL in the case of the window size of $15.87\boldsymbol{\tau_0}$ at lower noise level $D$. Whereas for larger noise levels $D$, higher MC provides a slight advantage for the signal prediction.